\documentclass[sigconf]{acmart}
\usepackage{appendix}
\usepackage[most]{tcolorbox}
\usepackage{soul}
\usepackage{multirow}
\usepackage{graphicx}
\usepackage{subcaption}
\usepackage{accents}
\usepackage{titlesec}
\usepackage{algorithm}
\usepackage{algpseudocode}
\usepackage{amsmath}
\usepackage{amsthm}
\usepackage{enumitem} 
\usepackage{bm}

\def\eg{{e.g., }}
\def\etal{{\emph{et al.}}}

\def\ie{\emph{i.e., }}

\newlist{myitemize}{itemize}{3}
\setlist[myitemize,1]{label=$\circ$,leftmargin=2.8ex}
\setlist[myitemize,2]{label=$\bullet$,leftmargin=2.8ex}
\setlist[myitemize,3]{label=$\diamond$, leftmargin=2.2ex}

\newenvironment{mbi}{\begin{myitemize}
        \setlength{\topsep}{0.6ex}\setlength{\itemsep}{0.16ex}\vspace{0.36ex}}
        {\end{myitemize}\vspace{0.16ex}}

\newcommand{\mei}{\end{myitemize}\vspace{0.6ex}}

\AtBeginDocument{%
  }

\copyrightyear{2025}
\acmYear{2025}
\setcopyright{acmlicensed}\acmConference[KDD '25]{Proceedings of the 31st ACM SIGKDD Conference on Knowledge Discovery and Data Mining V.2}{August 3--7, 2025}{Toronto, ON, Canada}
\acmDOI{XX.XXXX/XXXXXXX.XXXXXXX}
\acmISBN{XXX-X-XXXX-XXXX-X/XXXX/XX}

\begin{document}

\title[Urban Region Pre-training and Prompting: A Graph-based Approach]{Urban Region Pre-training and Prompting: \\A Graph-based Approach}
\author{Jiahui Jin}
\email{jjin@seu.edu.cn}

\affiliation{%
  \institution{School of Computer Science and
 Engineering, Southeast University}
  \city{Nanjing}
  \country{China}
}

\author{Yifan Song}
\email{yifansong2000@seu.edu.cn}
\affiliation{%
  \institution{School of Computer Science and
 Engineering, Southeast University}
  \city{Nanjing}
  \country{China}}

\author{Dong Kan}
\email{dongkan@seu.edu.cn}
\affiliation{%
  \institution{School of Computer Science and
 Engineering, Southeast University}
  \city{Nanjing}
  \country{China}
}

\author{Haojia Zhu}
\email{zhuhaojia@seu.edu.cn}
\affiliation{%
  \institution{School of Computer Science and
 Engineering, Southeast University}
  \city{Nanjing}
  \country{China}
}

\author{Xiangguo Sun}
\email{xiangguosun@cuhk.edu.hk}
\affiliation{%
  \institution{Department of Systems Engineering and Engineering Management, The~Chinese~University~of~Hong~Kong}
  \city{Hong Kong}
  \country{China}}

\author{Zhicheng Li}
\email{lizhicheng@seu.edu.cn}
\affiliation{%
  \institution{School of Computer Science and
 Engineering, Southeast University}
  \city{Nanjing}
  \country{China}
}

\author{Xigang Sun}
\email{xigangsun@seu.edu.cn}
\affiliation{%
  \institution{School of Computer Science and
 Engineering, Southeast University}
  \city{Nanjing}
  \country{China}
}

\author{Jinghui Zhang}
\email{jhzhang@seu.edu.cn}
\affiliation{
  \institution{School of Computer Science and
 Engineering, Southeast University}
  \city{Nanjing}
  \country{China}
}

\renewcommand{\shortauthors}{Jiahui Jin et al.}

\begin{abstract}
Urban region representation is crucial for various urban downstream tasks. However, despite the proliferation of methods and their success, acquiring general urban region knowledge and adapting to different tasks remains challenging. Existing work pays limited attention to the fine-grained functional layout semantics in urban regions, limiting their ability to capture transferable knowledge across regions. Further, inadequate handling of the unique features and relationships required for different downstream tasks may also hinder effective task adaptation. In this paper, we propose a \textbf{G}raph-based \textbf{U}rban \textbf{R}egion \textbf{P}re-training and \textbf{P}rompting framework (\textbf{GURPP}) for region representation learning. Specifically, we first construct an urban region graph and develop a subgraph-centric urban region pre-training model to capture the heterogeneous and transferable patterns of entity interactions. This model pre-trains knowledge-rich region embeddings using contrastive learning and multi-view learning methods. To further refine these representations, we design two graph-based prompting methods: a manually-defined prompt to incorporate explicit task knowledge and a task-learnable prompt to discover hidden knowledge, which enhances the adaptability of these embeddings to different tasks.  Extensive experiments on various urban region prediction tasks and different cities demonstrate the superior performance of our framework.
\end{abstract}



\maketitle
\section{Introduction}
An urban region is a geographical area comprising diverse spatial entities (\eg~shopping centers and roads) and their interactions (\eg~nearby and located in). Recently, urban region representation learning, which aims to extract vector embeddings from these entities, has become a significant research focus~\cite{zhang2024towards}. Effective urban region representations can be applied to various tasks, such as region popularity prediction~\cite{li2024urban, fu2019efficient, zhou2023heterogeneous, wu2022multi}, house price prediction~\cite{wang2017region}, and crime prediction~\cite{wang2017region, zhou2023heterogeneous, wu2022multi, zhang2021multi}, enabling urban planners and businesses to enhance decision-making processes.  

\begin{figure}[t]
    \centering\includegraphics[width=\linewidth]{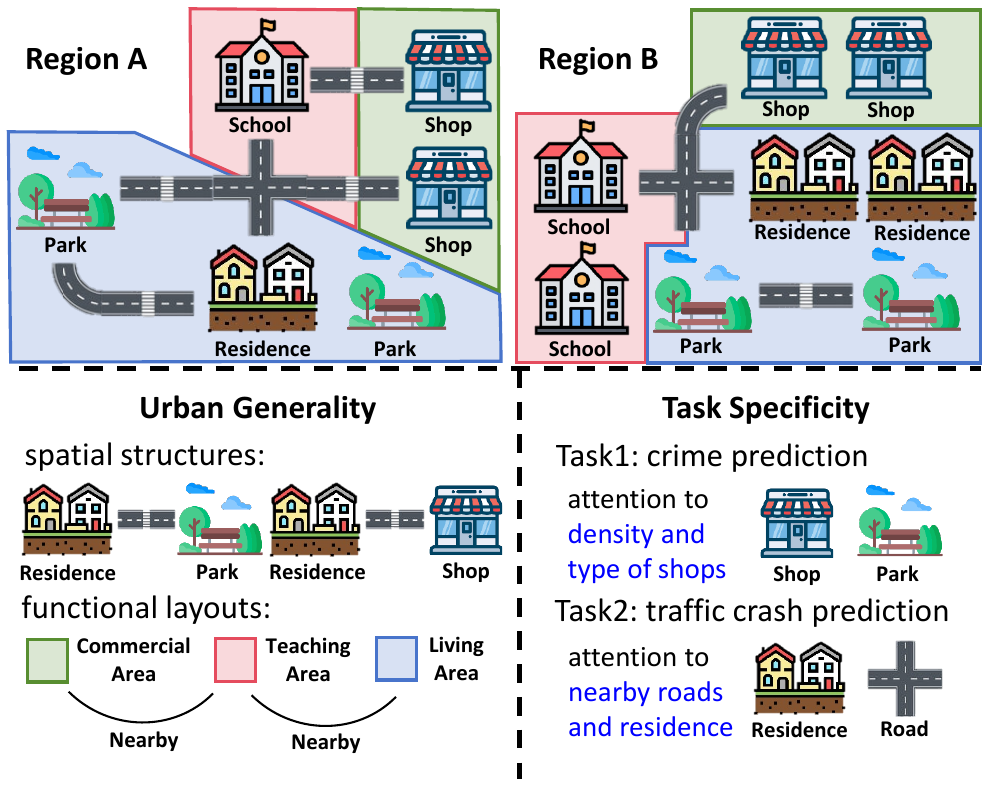}
    \vspace{-0.3cm}
    \caption{Example of urban generality and task specificity.} \label{fig:example}
    \vspace{-0.4cm}
\end{figure}

However, the complexity of urban environments and the diversity of applications pose significant challenges.
\textbf{On the one hand, urban regions often appear similar in their spatial structures and functional layouts}, because the interactions among entities within these regions follow similar patterns. These shared patterns serve as generic knowledge across urban areas, facilitating the development of generalizable region representation models. As illustrated in Figure \ref{fig:example}, Regions A and B showcase typical urban regions. In these examples, spatial structures are demonstrated by the proximity of residences to parks and shops, which enhances community convenience and accessibility. Functional layouts are characterized by the clustering of similar entities within areas, such as shops in commercial areas and schools in teaching areas, and the close proximity of these distinct areas to each other. Therefore, an effective urban region representation model needs to capture the heterogeneous and generic patterns of interactions among entities within regions. \textbf{On the other hand, despite the broad similarities, each urban region and downstream task possesses unique details.} Specific tasks may require attention to different features within urban data. For example, in traffic crash prediction, the density of roads and the number of residents living nearby might be critical features, as they directly influence the probability of accidents occurring. Meanwhile, crime prediction models might focus on the density and type of POIs (Point-of-Interests) within the region, such as bars or late-night convenience stores. As a consequence, the balance between generality and specificity raises a crucial question: \ul{\emph{Can we design an urban region representation framework that effectively integrates generalizable knowledge while accommodating the specific needs of distinct tasks?}}

To solve this problem, this paper presents a simple yet effective framework with pre-training and prompting on geographical data, since this new paradigm has shown remarkable success in balancing embedding generality and task specificity in the areas like natural language processing~\cite{liu2023pre, roberts2020much}, computer vision~\cite{zhou2022learning, zhou2022conditional}, and graphs~\cite{sun2023all,
sun2023Graph-Prompt-Learning}. The intrinsic reason behind this superiority, as many researchers reported~\cite{wang2024does}, lies in the powerful capability of data manipulation from the prompt side, and knowledge learning from the pre-training side, which can offer flexible adaptation for various downstream tasks and useful knowledge for generalizable representation. However, handling such complex non-linear geographical data requires prior experience. Challenges are twofold:

\begin{mbi}
    \vspace{-4pt}
    \item \textbf{Difficulties in capturing the heterogeneous and generic knowledge from complex spatial entity interactions.} Existing methods mostly predefine aggregations and statistics of some urban entities and relationships, and then use these these predefined features for downstream tasks~\cite{yang2022classifying,zhang2022region,li2023urban}, which often fail to capture the spatial structures and functional layouts between various entities within regions. 
    \item \textbf{Difficulties in capturing task-specific patterns.} Existing methods usually use simple prompts like directly concatenating them to the pre-trained embeddings~\cite{yan2024urbanclip,zhou2023heterogeneous}, which might work for certain domains but might fall short for urban region tasks with heterogeneous and complex geographical entity interactions.
    \vspace{-4pt}
\end{mbi}

\begin{figure}[t]
    \centering
    \includegraphics[width=.95\linewidth]{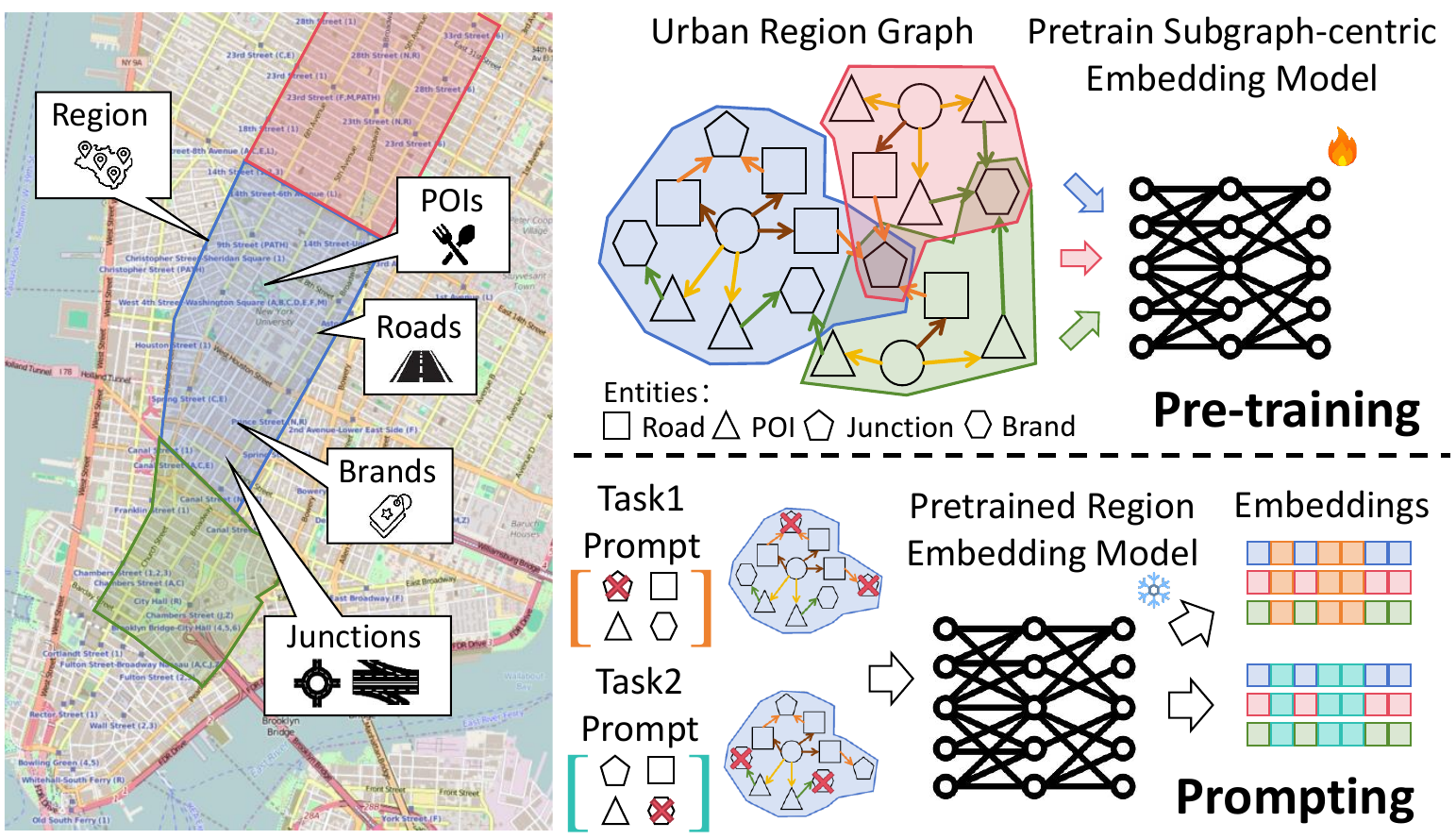}
    \vspace{-0.2cm}
    \caption{Main idea of urban pre-training and prompting.}
    \vspace{-0.3cm}
    \label{fig:intro_framework}
\end{figure}

In response to these challenges, we propose GURPP, a \textbf{G}raph-based \textbf{U}rban \textbf{R}egion \textbf{P}re-training and \textbf{P}rompting Framework as shown in Figure \ref{fig:intro_framework}. Specifically, GURPP leverages the expressive power of urban graphs to capture both generic and task-specific urban features. By constructing the urban region graph, GURPP effectively extracts the common spatial structures and functional layouts inherent to urban areas. Simultaneously, GURPP addresses the specificity required for different downstream tasks through a two-stage process. First, a subgraph-centric pre-training model, enhanced by multi-view self-supervised learning, captures diverse urban patterns—ranging from mobility trends and region imagery to intricate spatial and functional relationships. Then, two complementary graph-based prompting strategies further refine these representations: a manually-designed prompt that incorporates explicit task-specific semantics by emphasizing critical relationships and entities, and a task-learnable prompt that adapts to the hidden complexities of urban data. This design aims to bridge the gap between general urban knowledge and task-specific requirements for urban region representation. Contributions are as follows.
\begin{mbi}
    \vspace{-4pt}
    \item \textbf{Graph-based Urban Region Framework:} We introduce GURPP, a novel graph-based pre-training \& prompting framework for urban region representation, which unifies the modeling of complex spatial entity semantics. This design enables the extraction of heterogeneous and generic urban features, overcoming the limitations of methods that rely on predefined aggregations.
    \item \textbf{Subgraph-centric Pre-training and Dual Prompting Models:} We develop a novel subgraph-centric pre-training model, integrated with multi-view self-supervised learning, to capture urban generic knowledge from various modalities. Additionally, we propose two graph-based prompting methods—a manually-designed prompt and a task-learnable prompt—to inject explicit and implicit task-specific signals into the representations, thereby enhancing adaptability to diverse urban prediction tasks.
    \item \textbf{Extensive Empirical Validation:} We demonstrate the effectiveness of GURPP through extensive experiments on six urban region prediction tasks across two cities. Our results show significant performance improvements, with an average increase of 13.57\% on MAE and 16.24\% on RMSE, highlighting the model's strong capabilities in real-world urban prediction tasks.
    \vspace{-4pt}
\end{mbi}
\begin{figure*}[t]
    \centering
\includegraphics[width=.89\linewidth]{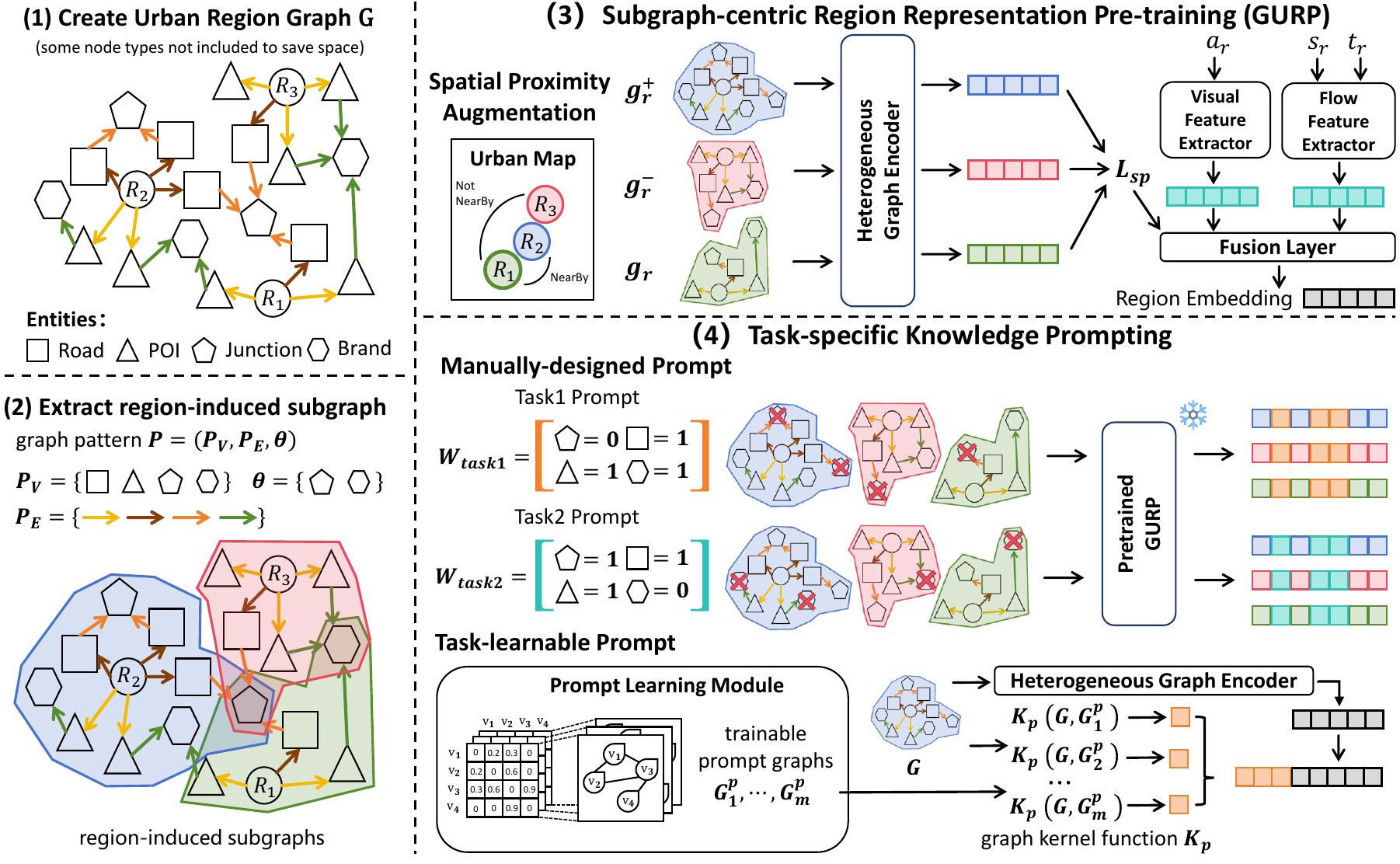}
    \vspace{-5pt}
    \caption{Overview of GURPP framework. (1) Construct the urban region graph that integrates specific urban entities and semantic relationships. (2) Extract a subgraph of each region according to the graph pattern. (3) Learn the subgraph-enhanced region representations with multi-view fusion. (4) Adapt the region embeddings with a manually-designed prompt method and a task-learnable prompt learning method.}
    \label{fig:Framework Overview}
\end{figure*} 

\section{GURPP Framework}\label{section:framework}
\subsection{Preliminaries}
\ul{\emph{Urban region.}} A city is divided into a set $\mathcal{R}$ of non-overlapping regions based on defined criteria, such as administrative boundaries~\cite{zhang2021multi, wu2022multi, liu2023urbankg, ning2024uukg}~and grid partitioning~\cite{jin2024learning, yuan2024unist, liu2019contextualized, feng2019finding}. Each region $r \in \mathcal{R}$ consists of multiple spatial entities, including points of interest (POIs), roads, and junctions. 

\noindent
\ul{\emph{Urban region representation learning.}} The urban region representation learning problem involves learning a mapping function $\zeta : \mathcal{R} \rightarrow \mathbb{R}^d$, which generates a low-dimensional embedding $\vec{h} \in \mathbb{R}^d$ for each region $r \in \mathcal{R}$. This mapping considers both the spatial structures and functional layouts of the spatial entities within $r$, with $d$ being the size of the embedding.

Once the mapping function $\zeta(\cdot)$ has been learned, it can be applied to various downstream prediction tasks, such as crime and check-in predictions. Let the task dataset be $\{(r_i, y_i)\}_{i=1}^{|\mathcal{R}|}$, where $y_i \in \mathbb{R}$ is a numerical indicator for region $r_i$. The downstream prediction tasks aim to predict $y_i$ based on the region embedding $\zeta(r_i)$ using a simple regression model like a Ridge Regression~\cite{hoerl1970ridge}. However, learning effective region representations for specific tasks is challenging. First, task datasets are often very limited. This means the function $\zeta(\cdot)$ must work well in few-shot or even zero-shot situations. Second, urban spatial entities are diverse and large in number. This requires $\zeta(\cdot)$ to pull out general knowledge while still focusing on features specific to the task.
\subsection{Graph-based pre-training and prompting}

We propose GURPP, a unified framework that integrates urban graph pre-training with prompt learning. We formalize the GURPP framework in Eq.~\eqref{eq:framework}, where GURPP organizes spatial entities and inter-regional relationships within regions using a graph construction function $G(\cdot)$: During the pre-training stage $\mathcal{M}^G_{pre}$, the graph structure enables fine-grained feature extraction, such as capturing granular details like store brands within regions; In the prompting phase $\mathcal{M}^G_{pmt}$, GURPP dynamically assembles spatial entities and relationships for a region $r$ via graph prompts $\mathcal{P}$ to generate task-specific representations. 
\begin{equation}
\zeta(\mathcal{P}, r; \mathcal{R}) = \mathcal{M}^G_{pre}\left(\mathcal{M}^G_{pmt}\left(\mathcal{P}, G(r)\right),  G(\mathcal{R})\right) \label{eq:framework}
\end{equation} 

GURPP integrates generalizable and task-specific knowledge within a unified framework. Its effectiveness can be demonstrated by recent advancements in graph prompt learning~{ \cite{wang2024does,yu2024generalized,li2024graph}}.  Moreover, GURPP offers flexibility by supporting versatile forms of prompting for the same pre-trained model, such as manually-designed prompting (\eg graph patterns~{ \cite{dong2017metapath2vec,huang2020graph}} and graph rules~{ \cite{fan2022discovering,fan2022parallel}}) or task-learnable prompting (\eg~soft prompting~{ \cite{liu2024can,chen2023dipping}}), as long as the prompt operation conforms to the urban graph schema.

In the following sections,  we detail the GURPP framework with a four-step workflow in Figure~\ref{fig:Framework Overview}. First, an urban graph is built to unify heterogeneous spatial entities (Section \ref{section:graph-construction}). Second, region-specific subgraphs are extracted using structural and interaction-based patterns to identify region-specific patterns and relationships within urban data (Section \ref{section:embedding}). Third, a pre-trained multi-view learning model refines these subgraphs into generalizable embeddings (Section \ref{section:multi-view learning}). Finally, task-specific knowledge is injected via prompt-tuning strategies (Section \ref{section:prompt}), enabling efficient adaptation without compromising the pre-trained backbone. 
\section{Pre-train General Representation} \label{section:pretrain}

In this section, we introduce the processes of constructing the urban region graph and pre-training subgraph-centric region embeddings using a multi-view learning approach. 

\subsection{Urban Region Graph Construction} \label{section:graph-construction}
An urban region graph models the spatial structures and functional layouts of urban regions. Given a set of regions $\mathcal{R}$, the urban region graph, denoted as $G(\mathcal{R}) = (V, E, T_V, T_E)$ (we simplify $G(\mathcal{R})$ to $G$ in the following), is a heterogeneous graph that captures multiple types of nodes and relationships. Here, $v \in V$ and $e \in E$ represent a node and an edge, respectively, while $\phi(v) \in T_V$ and $\psi(e) \in T_E$ are type functions that assign a specific type to a node or edge. 

The urban region graph follows a schema that defines the relationships among various types of nodes and edges (see in Appendix~\ref{app:graph construct}). Without loss of generality, we consider three types of spatial entities: POIs, roads, and junctions. 
Each spatial entity carries rich semantics; for instance, POIs have categories like shops and parks, and brands such as Dunkin' and Starbucks. Roads include types such as highways and streets and junctions cover categories like roundabouts. 
This design effectively captures spatial structures and functional layouts at a granular level. For instance, two nearby regions containing coffee shops with brands of Blue Bottle and Dunkin' are treated the same in existing statistic-based methods due to identical POI categories. In contrast, our region graph leverages semantic relations like BrandOf to distinguish them, enabling fine-grain region understanding, such as identifying high-end vs. casual commercial areas. 

We construct the urban region graph from POI datasets and online maps. Since the graph contains heterogeneous nodes and edges, we first adopt the knowledge graph embedding technique TransR~\cite{lin2015learning}~to initialize the representations of all nodes and edges, aligning the heterogeneous information into a unified vector space.

\subsection{Subgraph-centric Region Embedding} \label{section:embedding}
We introduce a subgraph-centric region embedding model to encode the spatial structures and functional layouts of a region $r$.

\subsubsection{Extract region-induced subgraph.}
The region-induced subgraph is a subgraph of $G$ that includes the region node $r$ and matches the graph pattern $\mathcal{P}=(P_V, P_E, \theta)$. Here, $P_V \subseteq T_V$ and $P_E \subseteq T_E$ are sets of node and edge types, respectively, which filter entities and relationships specifically useful for region embeddings.  The set $\theta \subseteq T_V$ indicates the node types that serve as termination points for subgraph extraction. As a result, the schema of the region-induced subgraph is a subset of the schema of the urban region graph. During the pre-training phase, we select all node and edge types and set $P_V=T_V, P_E = T_E$, \ie $\mathcal{P}=(T_V,T_E,\theta)$ with the aim of comprehensively characterising the rich information of a region and thereby learning general knowledge. The region-induced subgraph $g_r = (V_r, E_r, P_V, P_E)$ includes nodes $V_r$ and edges $E_r$ that can be reached from $r$ and whose types match $P_V$ and $P_E$. We extract  $g_r$ recursively, starting with $V_r = \{r\}$ and $E_r = \varnothing$ and repeatedly expanding $V_r$ and $E_r$ until the termination points are reached. Here, the maximum boundary for subgraph extraction is set to the two-hop neighbors of the region node according to the graph schema.

\vspace{-0.2cm}
\subsubsection{Region subgraph encoding.}  In Section~\ref{section:graph-construction}, each node in the subgraph $ g_r $ has been assigned a global initial feature vector $\vec{v}$ using TransR. We then encode each subgraph $ g_r $ into a vector $\vec{h}_r$ to further incorporate the local features of region $ r $. Specifically, we use a heterogeneous graph encoder $\textsf{gEnc}(\cdot)$ to extract the heterogeneous and generic patterns within $g_r$, ultimately outputting the contextualized embedding of the subgraph.

We choose the Heterogeneous Graph Transformer~(HGT)~\cite{hu2020heterogeneous} as the subgraph node encoder $\textsf{gEnc}(\cdot)$ due to its capability of capturing node and edge-type semantics. By using HGT, $\textsf{gEnc}(\cdot)$ characterizes the heterogeneity of input graph by maintaining a set of node and edge-type dependent parameters, and the output embedding of each node $\vec{v}^{HGT}$ that integrates the global initial embeddings generated by TransR, and node and edge-type semantics of the subgraph $g_r$. This trait aligns with our need to capture the semantic features of heterogeneous entities and relationships within each region-induced subgraph.

Obtaining the representations of each node in the subgraph, we further derive the graph-level representation of $g_r$ through two steps of aggregation. For each subgraph $g_r$, we first use $\text{SUM}(\cdot)$ function to aggregate the representations of nodes of the same type to integrate the impact of the number of entities on the representation, obtaining the node-type  embedding $\vec{\tau}_i$:
$\vec{\tau}_i = \sum_{\phi(v)=\tau_i} \vec{v}^{HGT}$, $\tau_i\in T_V, \forall v \in V_r$. Then we concatenate all representations of node types, and encode them with a Linear layer to capture the interaction between various types of nodes: $\vec{h}_r = \text{Linear}\left(\text{CONCAT}_{|T_V|}(\vec{\tau}_i)\right).$
The output $\vec{h}_r$ of $\textsf{gEnc}(\cdot)$ is the final embedding of region-induced subgraph $g_r$. Once the graph encoder is pretrained, any graph that conforms to the pattern $\mathcal{P}$ can be encoded by $\textsf{gEnc}(\cdot)$ to generate the corresponding representations, demonstrating high flexibility.

\vspace{-0.1cm}
\subsection{Multi-view Self-supervised Learning}
\label{section:multi-view learning}
We learn region representations in a self-supervised manner by integrating multiple views such as spatial proximity, region imagery and mobility patterns to characterize general region features.

\vspace{-0.1cm}
\subsubsection{Spatial proximity.} 
The First Law of Geography~\cite{tobler1970computer} emphasizes the concept of spatial proximity, stating that ``\textit{everything is related to everything else, but near things are more related than distant things}'', which means two regions with spatial proximity should exhibit greater similarity in their representation.

We utilize a triplet network~\cite{hoffer2015deep}~to capture region similarity based on spatial proximity. For a given subgraph $ g_r $ of region $r$, which serves as the anchor sample, we construct its positive and negative samples based on spatial proximity. If region $r'$ is adjacent to $r$ (\ie~there is a ``NearBy'' type of edge connecting the corresponding nodes), we take the subgraph of $ r' $ as the positive sample, denoted by $g_r^+$. Otherwise, we take the subgraph as the negative sample, denoted by $g_r^-$. We input $g_r$, $g_r^+$, and $g_r^-$ to the graph encoder $\textsf{gEnc}(\cdot)$, obtaining the vector representations $\vec{h}_r$, $\vec{h}_r^{+}$, and $\vec{h}_r^{-} \in \mathbb{R}^d$, respectively. Then we use triplet loss to minimize the distance between adjacent regions and maximize it between distant regions, where $\delta$ represents the margin parameter of triplet loss:
\begin{equation*}
    \mathcal{L}_{sp} = \sum_{r\in\mathcal{R}} \max\{||\vec{h}_r-\vec{h}_r^{+}||_2-||\vec{h}_r-\vec{h}_r^{-}||_2+\delta,0\}.
\end{equation*}

\vspace{-4pt}
\subsubsection{Region imagery.} Urban imagery encompasses various types of image information that describe the visual characteristics of a city, such as satellite images, street view images, and more. In this paper, we use satellite images as the source of urban imagery. For a given region $r$, its imagery can be represented as the set $\{a_1, a_2, \cdots, a_m\}$. We employ a pre-trained visual model to extract visual features of the region $\vec{h}_r^{img}$ and fuse these features with the region's representation with the contrastive loss $\mathcal{L}_{img}$ (see Appendix~\ref{app:imagery} for details).

\subsubsection{Mobility pattern.}
Mobility pattern describes the dynamics of urban systems and is typically represented by trajectory data. For a given region $ r $, the outflow feature $\mathbf{s}_r$ and inflow feature $\mathbf{t}_r$ are defined by calculating the number of outgoing and incoming trips within each time period. This results in the outflow and inflow feature sets $\{\mathbf{s}_1, \mathbf{s}_2, \cdots, \mathbf{s}_N\}$ and $\{\mathbf{t}_1, \mathbf{t}_2, \cdots, \mathbf{t}_N\}$, respectively, where $\mathbf{s}_i, \mathbf{t}_i \in \mathbb{R}^l$, and $ l $ is the number of intervals (e.g., 24). We encode the inflow and outflow features with an encoder (such as a Multi-Layer Perceptron) to obtain $d$-dimensional embeddings: $\vec{h}^{\text{src}}_r = \textsf{Encoder}(\mathbf{s}_r)$ and $ \vec{h}^{\text{dst}}_r = \textsf{Encoder}(\mathbf{t}_r).$ Here we adopt the reconstruction loss $\mathcal{L}_{flow}$ to optimize the two embeddings (see Appendix~\ref{app:flow} for details).

\subsubsection{Self-supervised region representation learning.}
We aim to jointly learn region representations from spatial proximity, imagery, and flow views to extract general urban region knowledge and embed this into a unified representation space. To achieve this, we propose a view-fusion task that encodes the region embedding $\vec{h}_r$, the imagery embedding $\vec{h}_r^{img}$, and the flow embeddings $\vec{h}_r^{src}$ and $\vec{h}_r^{dst}$ into a combined vector $\hat{h}_r$. We then decode $\hat{h}_r$ using separate decoders to reconstruct the original embeddings.

To implement this idea, we employ a fusion layer to integrate the embeddings: $\hat{h}_r = \text{ReLU}(\textbf{W}[\vec{h}_r, \vec{h}_r^{img}, \vec{h}_r^{src}, \vec{h}_r^{dst}] + \textbf{b})$,
where $[\cdot]$ denotes the concatenation operation, and $\textbf{W}$ and $\textbf{b}$ are trainable weights. We design a multi-view prediction task to train the fusion layer with the loss function $\mathcal{L}_{fuse}$, defined by the ability to reconstruct the representation of each individual embedding $\vec{h}_r, \vec{h}_r^{img}, \vec{h}_r^{src}, \vec{h}_r^{dst}$ from the fused representation $\hat{h}_r$:
\[
\mathcal{L}_{fuse} = \sum_{r \in \mathcal{R}} \sum_{k=1}^4 \|\vec{h}^k_r - \textsf{Decoder}^k(\hat{h}_r)\|_2^2.
\]
Here, $\vec{h}^k_r$ ($k=1,2,3,4$) represents one of the four embeddings, and $\textsf{Decoder}^k$ is the corresponding decoder (implemented by a linear layer). This fusion module is easy to be extended to more views. The final loss function can be formulated as follows:
\[
\mathcal{L} = \mathcal{L}_{sp} + \mathcal{L}_{img} + \mathcal{L}_{flow} + \mu \mathcal{L}_{fuse}.
\]
where $\mu$ is a parameter that balances the weight of the fusion loss.
\section{Prompt with Task-specific Knowledge} \label{section:prompt}
We propose two graph-based prompt tuning methods, a manually-designed prompt method and a task-learnable prompt learning method, to extract explicit and implicit task-specific knowledge.

\subsection{Manually-designed Prompt} \label{section:manual-prompt}
\begin{figure}[htbp]
    \centering
    \vspace{-0.4cm}
    \includegraphics[width=1.1\linewidth]{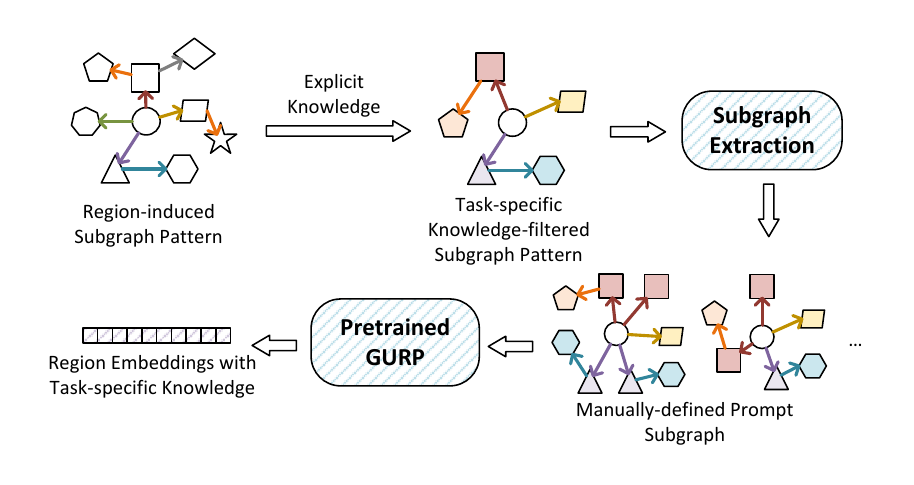}
    \vspace{-0.8cm}
    \caption{Manually-designed prompt.}
    \label{fig:manually-designed prompt}
    \vspace{-0.2cm}
\end{figure}

In the manually-designed prompt, we adjust the subgraph structure, \ie defining the graph pattern (\ie prompt) $\mathcal{P}(P_V, P_E, \theta)$ in Section~\ref{section:embedding}, to emphasize specific relationships and entities relevant to the target task, as shown in Figure~\ref{fig:manually-designed prompt}.

We enhance this process by manually adjusting the proportion of different types of edges and entities in the subgraph, thereby influencing the resultant embeddings. To achieve this, we first define a set of task-specific entity weights $\mathbf{W}_{task}$ that reflect the importance of various types of entities for the downstream task. These weights are designed based on the domain knowledge and specific requirements of task. Formally, we define $\mathbf{W}_{task} = \{ w_{\tau_i} \mid \tau_i \in P_V \},$ where $P_V$ is the set of all possible node types in region-induced subgraph, and $w_{\tau_i}\in[0,1]$ represents the importance of entity type $\tau_i$. The weights $w_{\tau_i}$ are manually defined and used as the deletion proportion for each entity type. We define a function $\textsf{Adjust}(\cdot)$ to delete nodes based on $\mathbf{W}_{task}$:~$\textsf{Adjust}(g_r, \mathbf{W}_{task}) \rightarrow g'_r$, where $g_r = (V_r, E_r)$ is the original subgraph. The adjusted subgraph $g'_r = (V'_r, E'_r)$ is input to the pre-trained GURP to generate the task-specific embeddings. The function $\textsf{Adjust}$ obtains the adjusted subgraph $g'_r = (V'_r, E'_r)$ as follows:

\textbf{(1) Node Deletion}: For each entity type $\tau_i \in \mathcal{E}$, delete $w_{\tau_i} \cdot |V_{\tau_i}|$ nodes of type $\tau_i$ from $g_r$, where $|V_{\tau_i}|$ is the number of nodes of type $\tau_i$ in $g_r$, such that $V'_r = V_r \setminus \{ v \mid \phi(v) = \tau_i \text{ with probability } 1 - w_{\tau_i} \}$.

\textbf{(2) Edge Deletion}: After node deletion, delete edges connected to the removed nodes, such that $E'_r = \{ e \mid e = (u, v) \in E; u, v \in V'_r \}$.

The manually-designed prompt is the graph pattern $\mathcal{P}$ with the set of task-specific entity weights $\mathbf{W}_{task}$, which guide the $\textsf{Adjust}(\cdot)$ function in modifying the subgraph $g_r$ to $g'_r$. Incorporating such prompts, we can tailor the urban region representations to better capture task-relevant entities and relationships. 

\subsection{Task-learnable Prompt} \label{section:learnable-prompt}
Apart from leveraging explicitly designed task-specific knowledge prompts to enrich region representations, we can also learn prompts based on the task data and pre-trained representations to induce implicit knowledge within the task data, thereby enhancing region representations.

\begin{figure}[htbp]
    \centering
    \vspace{-0.4cm}
    \includegraphics[width=1.05\linewidth]{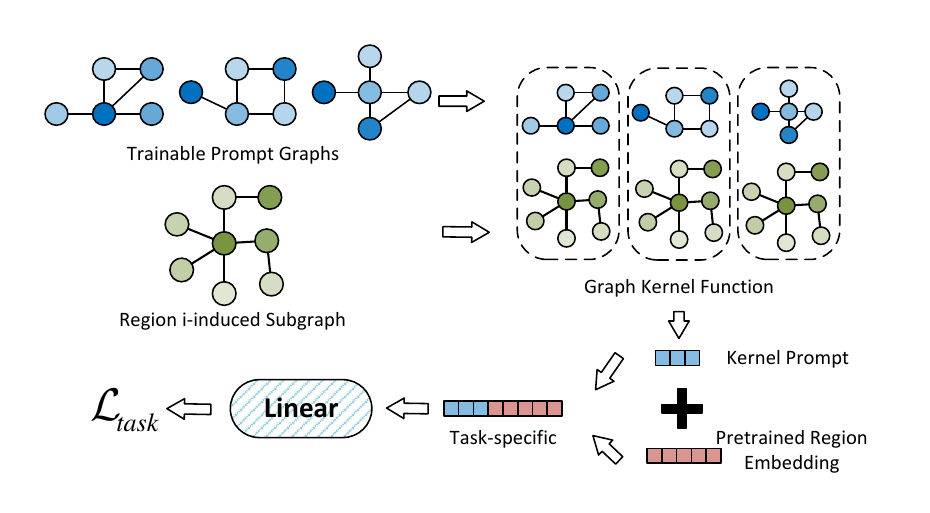}
    \vspace{-0.4cm}
    \caption{Task-learnable prompt.}
    \label{fig:task-learnable prompt}
    \vspace{-0.1cm}
\end{figure}

Considering that we describe the attributes of an urban region through subgraph induced with region nodes from $G$, it is intuitive to describe the task adaptation to different region attributes with graphs as well. Therefore, we propose a prompting module to use prompt graphs to portray the adaptation of downstream task relative to the original region embedding, as shown in Figure~\ref{fig:task-learnable prompt}. 

We introduce the graph kernel method~\cite{vishwanathan2010graph,feng2022kergnns} to measure the similarity between the prompt graphs and the region subgraphs due to its performance and computational efficiency. Let $\mathcal{P}$ denote the set (\ie prompt) consisting of $m$ trainable (hidden) prompt graphs, where each graph $p \in \mathcal{P}$ is defined as $p = (V^p, E^p, \mathbf{X}^p)$. Here, $\mathbf{X}^p \in \mathbb{R}^{|V^p| \times d}$ denotes the node attribute matrix, where $d$ is the attribute dimension. For each region subgraph $g_r$, we utilize an attribute graph kernel function $K_{p}(\cdot,\cdot)$ to measure the similarity between $g_r$ and each prompt graph $p_1, p_2, \ldots, p_m$. Specifically, we employ the $P$-step random walk kernel (more details in Appendix~\ref{app:kernel}) to compute the kernel value between region subgraph $g_r$ and each prompt graph $p_i$. These values are concatenated to form an $m$-dimensional prompt vector $\vec{h}_\mathcal{P} \in \mathbb{R}^m$ for each region. 

The prompt vector is used as a data operator to modify the pre-trained graph model. Conceptually, we can concatenate $\vec{h}_\mathcal{P}$ with each  initial node feature vector $\vec{v}$ and feed the result into a linear layer $\textsf{Linear}(\cdot)$, such that each entity is modified by $\vec{v}_* = \textsf{Linear}(\vec{h}_\mathcal{P} || \vec{v})$. To reduce the complexity, in practice, we concatenate $\vec{h}_\mathcal{P}$ with the pre-trained region embedding $\vec{h}_r$ instead of each entity embedding. To learn the prompting model, we update the prompt graphs and linear layer based on specific downstream task with a simple MSE loss function.  
\begin{table*}[t]
  \caption{Main performance comparison of different models on downstream tasks. The best results for each metric are highlighted in bold, the second-best are underlined, and the third-best are marked with an asterisk.}
  \vspace{-0.4cm}
  \label{tab:main_performance_result}
  \resizebox{\textwidth}{!}{
      \begin{tabular}{c|cc|cc|cc|cc|cc|cc|cc|cc}
        \toprule
        \multirow{3}[1]{*}{Model} & \multicolumn{8}{c|}{New York City~(NYC)} & \multicolumn{8}{c}{Chicago~(CHI)} \\
        \cmidrule{2-17}
         & \multicolumn{2}{c|}{Crime} & \multicolumn{2}{c|}{Check-in } & \multicolumn{2}{c|}{Carbon} & \multicolumn{2}{c|}{Income} & \multicolumn{2}{c|}{Crime} & \multicolumn{2}{c|}{Crash} & \multicolumn{2}{c|}{Carbon} & \multicolumn{2}{c}{Education}\\
        \cmidrule{2-17}
        &
        \multicolumn{1}{c}{MAE} & \multicolumn{1}{c|}{RMSE} & 
        \multicolumn{1}{c}{MAE} & \multicolumn{1}{c|}{RMSE} &
        \multicolumn{1}{c}{MAE} & \multicolumn{1}{c|}{RMSE} & \multicolumn{1}{c}{MAE} & \multicolumn{1}{c|}{RMSE} &
        \multicolumn{1}{c}{MAE} & \multicolumn{1}{c|}{RMSE} & \multicolumn{1}{c}{MAE} & \multicolumn{1}{c|}{RMSE} &
        \multicolumn{1}{c}{MAE} & \multicolumn{1}{c|}{RMSE} & \multicolumn{1}{c}{MAE} & \multicolumn{1}{c}{RMSE} \\
        \midrule
        TransR-N & 111.32 & 148.86 & 534.70 & 805.28 & 115.65 & 175.73 & 36799 & 45098 & 209.07 & 307.45 & 106.91 & 176.86 & 97.15 & 126.59 & 653.68 & 932.56 \\
        TransR-G & 109.22 & 146.91 & 530.43 & 802.43 & 116.80 & 179.25 & 36363 & 44571 & 208.40 & 306.16 & 104.68 & 174.42 & 95.76* & 124.37* & 601.97 & 875.24 \\
        \midrule
        node2vec & 138.46 & 175.80 & 573.12 & 837.44 & 174.45 & 225.42 & 45330 & 56182 & 246.07 & 349.97 & 121.90 & 174.91 & 115.62 & 148.95 & 832.89 & 1139.33 \\
        GAE & 96.85 & 126.36 & 349.94 & 542.92 & 115.99 & 174.35 & 35338 & 43200 & 200.61* & 300.24 & 100.54 & 173.32 & \underline{95.29} & \underline{123.66} & 702.82 & 986.70 \\
        \midrule
        MVURE & 72.47* & 99.84 & 310.97 & 498.08 & \underline{99.30} & \underline{151.01} & \underline{24031} & 35351 & 206.08 & 289.86* & 96.95 & 148.87* & 98.19 & 130.31 & 593.80 & 843.10 \\
        MGFN & 73.50 & 97.44* & 300.15 & 456.63* & 136.49 & 185.82 & 30266 & 40717 & 214.02 & 310.83 & 108.37 & 179.85 & 96.56 & 125.50 & 709.91 & 998.03 \\
        HRE & 76.93 & 102.07 & 329.55 & 513.80 & 102.95 & 160.58 & 25903 & 35418 & 206.61 & 298.39 & 99.95 & 158.63 & 97.11 & 133.00 & 555.56 & 806.60 \\
        HREP & 74.85 & 100.04 & 320.34 & 505.52 & 102.49 & 159.68 & 24643 & \underline{34148} & 206.59 & 298.24 & 99.77 & 158.66 & 97.19 & 132.91 & 555.01 & 806.36 \\
        ReCP & 73.85 & 100.56 & 292.87* & 524.89 & \textbf{95.55} & \textbf{150.84} & 24440* & 35044 & 207.49 & 298.97 & 93.30* & 157.52 & 95.84 & 125.73 & 455.98* & 708.37*\\
        \midrule
        GURP & \underline{72.09} & \underline{93.22} & \underline{271.69} & \underline{412.92} & 108.17 & 158.03 & 25268 & 34943* & \underline{161.27} & \underline{225.24} & \underline{80.36} & \underline{108.78} & 97.83 & 133.03 & \underline{455.43} & \underline{677.36}\\
        GURPP & \textbf{71.48} & \textbf{91.61} & \textbf{261.45} & \textbf{388.23} & 100.53* & 152.70* & \textbf{22902} & \textbf{31945} & \textbf{146.39} & \textbf{203.05} & \textbf{65.30} & \textbf{92.11} & \textbf{64.844} & \textbf{88.56} & \textbf{420.62} & \textbf{656.98} \\
        Enhance & 1.36\% & 5.98\% & 10.73\% & 14.98\% & -4.95\% & -1.21\% & 4.70\% & 6.45\% & 27.03\% & 29.95\% & 30.01\% & 38.12\% & 31.95\% & 28.39\% & 7.76\% & 7.26\% \\
        \bottomrule
    \end{tabular}
    }
\end{table*}
\section{Experiments}
\label{section:exp}

We conduct extensive experiments on six downstream tasks of two cities from the following perspectives: (1) overall performance comparison with SOTA baselines, (2) effectiveness analysis of two prompting methods, (3) few-shot and zero-shot generalization capabilities, (4) ablation studies, (5) scalability evaluation, and (6) comparison with language-based prompting methods.

\subsection{Experimental Setup}

\subsubsection{Datasets.} We collect the datasets of region division, POI, urban imagery, taxi trips, road network and six downstream tasks for two representative cities in the United States: New York City (NYC) and Chicago (CHI). 

\subsubsection{Baselines.}
We compare our model with the following representative baselines: knowledge graph embedding methods and its variants (\textbf{TransR}~\cite{lin2015learning}, \textbf{TransR-N}, \textbf{TransR-G}),  graph embedding methods (\textbf{node2vec}~\cite{grover2016node2vec}, \textbf{GAE}~\cite{kipf2016variational}), and urban region embedding methods (\textbf{MVURE}~\cite{zhang2021multi}, \textbf{MGFN}~\cite{wu2022multi}, \textbf{HREP}~\cite{zhou2023heterogeneous},\textbf{ReCP}~\cite{li2024urban}). We follow the descriptions in these works to preprocess the data into the corresponding format and run the baselines with the parameter settings specified in the papers to ensure fairness. 

\subsubsection{Metrics and Implementation.} We use three widely recognized evaluation metrics to evaluate the prediction performance: the coefficient of determination ($R^2$), root mean squared error (RMSE), and mean absolute error (MAE). Better performance is indicated by a higher $R^2$ and lower RMSE and MAE values.
Our model includes three version: 1) \textbf{GURP} represents the pre-train module, 2) $\textbf{GURPP}_M$ is GURP with manually-designed prompt, and 3) $\textbf{GURPP}_T$ is GURP with task-learnable prompt. 
Details on datasets and implementation are in Appendix \ref{app:experiment details}.

\subsection{Overall Performance}
We evaluate the overall performance of the baselines and our model on datasets collected from two cities, covering six different downstream tasks. The results of the experiments are shown in Table \ref{tab:main_performance_result}, from which we find the following.

(1) GURPP demonstrates superior performance across both cities and tasks. In the pre-training phase, GURP achieves state-of-the-art performance on 6 out of 8 tasks, with particularly significant improvements on CHI datasets (27.03\%-31.95\% MAE reduction on Crime/Crash prediction). When enhanced with prompting methods, GURPP further improves performance by an average of 11.30\% (MAE) and 10.16\% (RMSE) across all tasks compared to GURP. The consistent improvements across heterogeneous tasks (12.46\% average MAE gain and 14.62\% RMSE gain over baselines) validate the effectiveness of our pre-training and prompting framework.

(2) The results highlight our model's exceptional adaptability to different urban contexts. Specifically, GURPP achieves average improvements of 24.2\% (MAE) and 25.9\% (RMSE) in CHI versus 3.0\% (MAE) and 6.6\% (RMSE) in NYC. This may be due to the fact that the baselines tend to be effective only at extracting information specific to certain data types, but they do not consider spatial structure and functional layouts among urban entities. For example, 
existing baselines rely on the flow view for information extraction, but the sparse content in the CHI dataset significantly degrades downstream task performance. In contrast, GURPP performs excellently on both urban datasets, demonstrating its robustness and universality across diverse urban environments.

(3) To further validate the effectiveness of our subgraph-centric and multi-view fusion module, we conduct experiments comparing the performance of models with and without these modules. Specifically, the result that TransR-G outperforms TransR-N by an average of 1.27\% (MAE) indicates that the subgraph-centric method can preserve the structural and contextual information of regions, providing more effective urban region embeddings. The average 32.55\% (MAE) improvement for GURP over TransR-G is mainly due to the heterogeneous feature extraction, which improves the model's comprehensive understanding of urban features.

\subsection{Effectivness of Prompting Methods}

\subsubsection{Manually-designed Prompt Study.}

In the real world, crime rates are closely linked to specific POI categories, roads, and junctions, with locations like ATMs, bars, and parks being more crime-prone. Road density and junction complexity also impact crime. Regarding user check-ins, brand popularity plays a key role, with users often preferring well-known brands like Starbucks or McDonald's. Based on these observations, we design two patterns: P1 (ignoring POI brand information) and P2 (randomly sampling POI, road, and junction nodes with a 0.9 ratio) to evaluate the performance of manually-designed prompts across different tasks.

\begin{table}[ht]
  \small
  \caption{\protect\parbox{0.48\textwidth}{Performance of manually-designed prompt $\text{GURPP}_M$.}}
  \vspace{-0.4cm}
  \label{tab:manually_result}
  \begin{tabular}{c|ccc|ccc}
    \toprule
    \multirow{2}[1]{*}{Model} & \multicolumn{3}{c|}{Crime Prediction~(NYC)} & \multicolumn{3}{c}{Check-in Prediction~(NYC)}\\
    \cmidrule{2-7}
    &
    \multicolumn{1}{c}{MAE} & \multicolumn{1}{c}{RMSE} & \multicolumn{1}{c|}{$R^2$} &
    \multicolumn{1}{c}{MAE} & \multicolumn{1}{c}{RMSE} & \multicolumn{1}{c}{$R^2$} \\
    \midrule
    GURP & 72.094 & 93.216 & 0.603 & \underline{271.689} & \underline{412.918} & \underline{0.745}\\
    $\text{GURPP}_M$-P1 & \textbf{68.538} & \textbf{88.698} & \textbf{0.640} & 273.201 & 429.564 & 0.724\\
    $\text{GURPP}_M$-P2 & \underline{69.421} & \underline{91.834} & \underline{0.614} & \textbf{262.437} & \textbf{397.466} & \textbf{0.764}\\   
    \bottomrule
  \end{tabular}
  \vspace{-0.2cm}
\end{table}

Results in Table \ref{tab:manually_result} show that focusing on factors such as POI categories, roads, and junctions, closely related to criminal behavior, can improve crime prediction performance. Meanwhile, removing POI brand information shows a significant decrease in the performance of check-in prediction. This demonstrates that our manually-designed prompting method can improve the prediction performance of downstream tasks with explicit knowledge injection.

\vspace{-0.1cm}
\subsubsection{Task-learnable Prompt Study.}
We utilize GURP pre-trained embedding with the simple prefix prompt (proposed by HREP\cite{zhou2023heterogeneous} denoted as \textbf{GURP+Pre}), which directly concatenates hidden vector before pretrained embedding without modeling task-adaptation, and our task-learnable prompt for comparative analysis. Both of these prompt methods are data-driven and seek to improve the region representation for specific downstream tasks. We conduct experiments on crime prediction task in NYC and crash prediction task in CHI, respectively, and the results are shown in Table \ref{tab:task_prompt_result}. 

Results show that our task-learnable prompt outperforms pre-trained embeddings across both tasks, with improvements of 2.16\% ($R^2$) in crime prediction and 17.20\% ($R^2$) in crash prediction, indicating that the prompt is highly effective in providing tailored guidelines for different downstream tasks. In addition, $\text{GURPP}_T$ outperforms the prefix prompt in both tasks. This is because the prefix prompt only utilizes the pre-trained region representations for prompting, while our graph-based prompting method also incorporates the original region subgraphs to prompt, better capturing the correlations between task and region attributes.

\begin{table}[t]
  \small
  \caption{Performance of task-learnable prompt $\text{GURPP}_T$.}
  \vspace{-0.3cm}
  \label{tab:task_prompt_result}
  \begin{tabular}{c|ccc|ccc}
    \toprule
    \multirow{2}[1]{*}{Framework} & \multicolumn{3}{c|}{Crime Prediction~(NYC)} & \multicolumn{3}{c}{Crash Prediction~(CHI)}\\
    \cmidrule{2-7}
    &
    \multicolumn{1}{c}{MAE} & \multicolumn{1}{c}{RMSE} & \multicolumn{1}{c|}{$R^2$} &
    \multicolumn{1}{c}{MAE} & \multicolumn{1}{c}{RMSE} & \multicolumn{1}{c}{$R^2$} \\
    \midrule
    GURP & \underline{72.094} & \underline{93.216} & \underline{0.603} & 80.362  & 108.781 & 0.622 \\ 
    GURP+Pre & 73.758 & 93.971 & 0.596 & \underline{66.330} & \underline{93.772} & \underline{0.719} \\
    $\text{GURPP}_T$ & \textbf{71.481} & \textbf{91.612} & \textbf{0.616} & \textbf{65.297} & \textbf{92.114} & \textbf{0.729} \\ 
    \bottomrule
  \end{tabular}
  \vspace{-0.4cm}
\end{table}

\vspace{-0.1cm}
\subsubsection{Comparison of Two Prompting Methods.} We compare the performance of two proposed prompting methods across different cities and tasks to evaluate their effectiveness in various scenarios. 
In addition to the two prompting patterns P1 and P2, we also define two additional patterns: P3 (randomly samples POI with a weight of 0.9 and ignores road and junction category) and P4 (randomly samples POI with a weight of 0.7 and ignores road and junction), to further validate the effectiveness of different prompting methods. Results show in Table~\ref{tab:dif_prompt_performance_mini}.
\begin{table}[t]
  \small
  \caption{Comparison of two prompting methods. }
  \label{tab:dif_prompt_performance_mini}
  \vspace{-0.3cm}
  \resizebox{0.48\textwidth}{!}{
  \begin{tabular}{c|c|c|c|c|c|c}  
    \toprule
    \multirow{3}[1]{*}{Method} & \multicolumn{3}{c|}{New York City~(NYC)} & \multicolumn{3}{c}{Chicago~(CHI)} \\
    \cmidrule{2-7}
     & \multicolumn{1}{c|}{Check-in} & \multicolumn{1}{c|}{Carbon} & \multicolumn{1}{c|}{Income} & \multicolumn{1}{c|}{Crime} & \multicolumn{1}{c|}{Carbon} & \multicolumn{1}{c}{Education} \\
    \cmidrule{2-7}
    &
    \multicolumn{1}{c|}{RMSE} &
    \multicolumn{1}{c|}{$R^2$} &
    \multicolumn{1}{c|}{MAE} & \multicolumn{1}{c|}{RMSE} & \multicolumn{1}{c|}{$R^2$} & 
    \multicolumn{1}{c}{MAE}\\
    \midrule
    GURP & 412.92 & \underline{0.23} & 25268 & \underline{225.24} & \underline{-0.33} & \underline{455.43} \\
    \midrule 
    $\text{GURPP}_T$ & \underline{388.23} & 0.17 & \underline{24745} & \textbf{203.05} & \textbf{0.41} & \textbf{420.62} \\
    \midrule
    $\text{GURPP}_M$-P1 & 429.56 & -0.13 & 27839 & 234.26 & -0.35 & 465.81 \\
    $\text{GURPP}_M$-P2 & 397.47 & -0.22 & 26392 & 234.74 & -0.40 & 458.59 \\
    $\text{GURPP}_M$-P3 & 464.90 & \textbf{0.28} & 28170 & 256.53 & -0.66 & 486.19 \\
    $\text{GURPP}_M$-P4 & \textbf{340.25} & 0.16 & \textbf{22902} & 238.36 & -0.49 & 502.86 \\
    \bottomrule
\end{tabular}
} 
\vspace{-0.4cm}
\end{table}

From the results, we observe that when the pre-trained city region data is relatively sparse (e.g., in cases of CHI), task-learnable prompting methods, especially our $\text{GURPP}_T$, achieve better enhancement effects. Conversely, when the data is more abundant (e.g., in cases of NYC), manually-designed prompts with explicitly injected knowledge perform better. Moreover, different manually-designed prompts explicitly capture distinct types of knowledge, leading to varying performance across different tasks, which demonstrates the flexibility of our proposed graph-based prompting approach.

\subsection{Few-shot \& Zero-shot Performance}
We compare the performance of $\text{GURPP}_T$ and baselines in few-shot and zero-shot scenarios to validate the generalizability of pre-trained representations and superiority of prompting methods.

\subsubsection{Few-shot Prediction.} With the pre-trained region representations as input, we randomly sample 10\% of the region data as the training set to develop the prediction model, while the remaining 90\% is designated as the testing set, to evaluate the performance of $\text{GURPP}_T$ with other baselines in few-shot scenarios.

Figure~\ref{fig:few_shot} shows that $\text{GURPP}_T$ outperforms all baselines in the few-shot scenario. This is partly because the pre-trained representations extract generic regional features that exhibit better adaptability in low-resource settings, which, when combined with task-learnable prompts, can better accommodate few-shot prediction. Additionally, our prompting method shows effective transferability compared to HREP.
\vspace{-0.1cm}
\subsubsection{Zero-shot Prediction.} 
\begin{figure}[h!]
    \centering
    \vspace{-0.3cm}
    \begin{subfigure}[b]{0.235\textwidth}
        \centering
        \includegraphics[width=\textwidth]{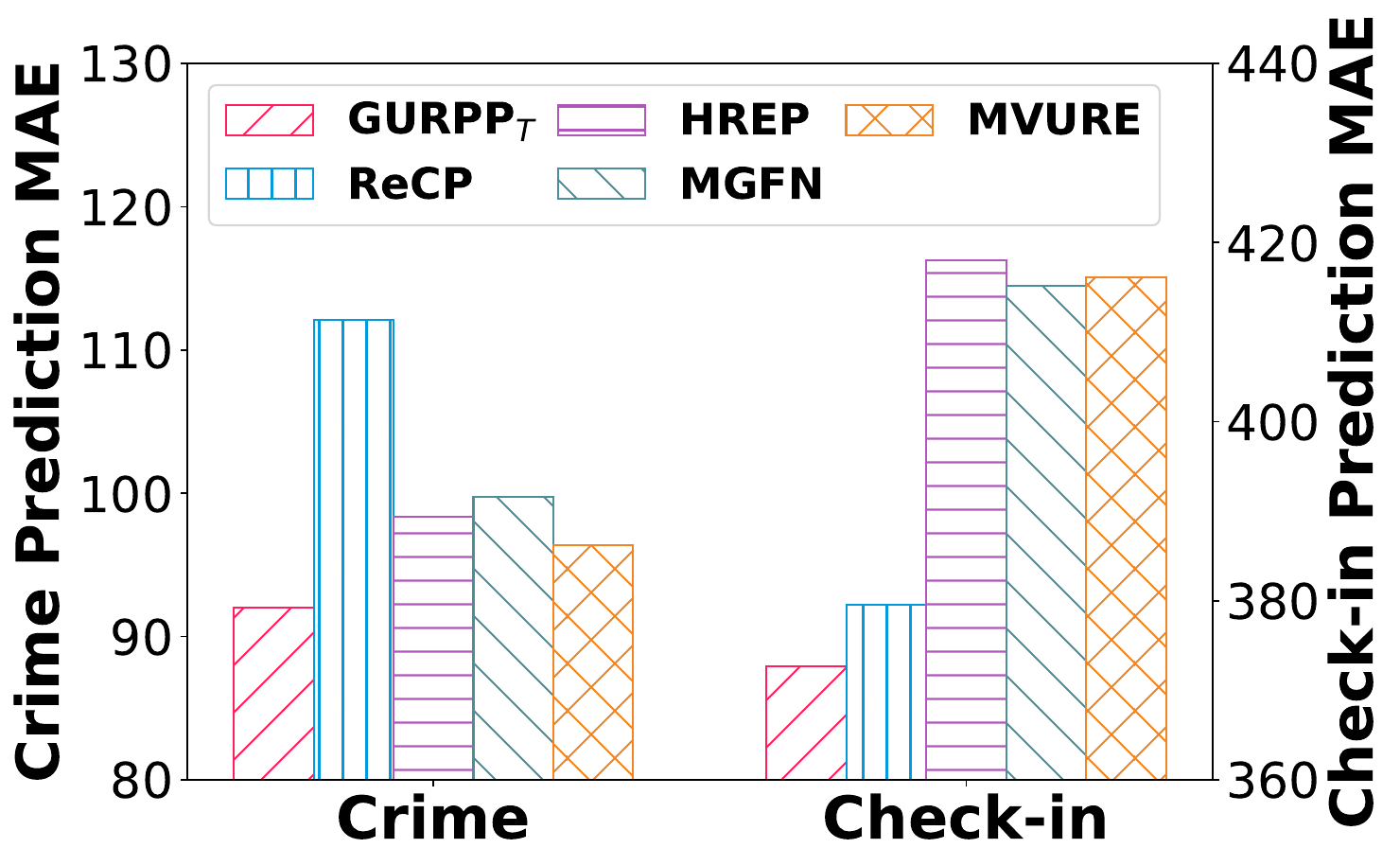}
        \caption{Few-shot prediction.}
        \label{fig:few_shot}
    \end{subfigure}
    \hfill
    \begin{subfigure}[b]{0.235\textwidth}
        \centering
        \includegraphics[width=\textwidth]{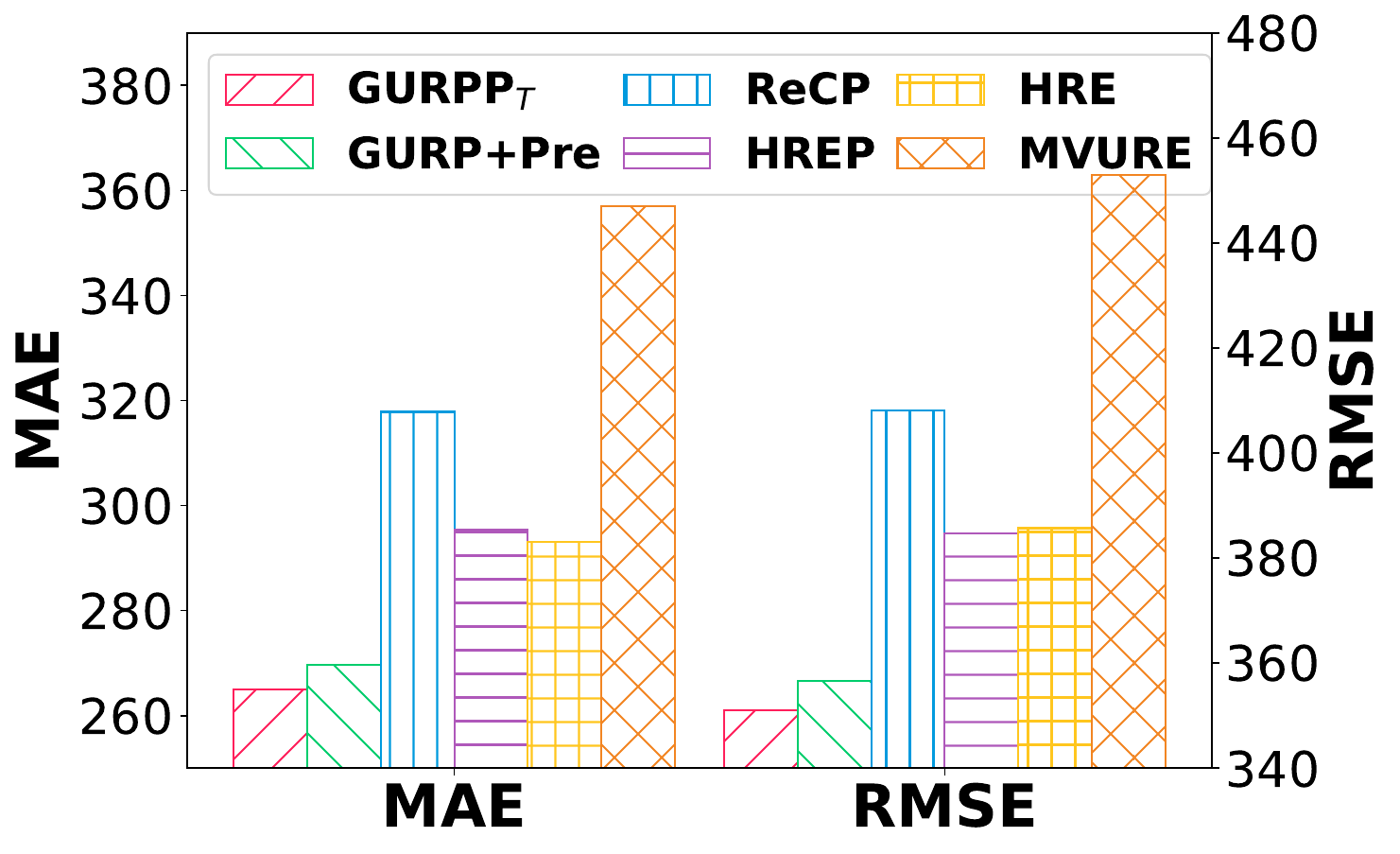}
        \caption{Zero-shot prediction.}
        \label{fig:zero_shot}
    \end{subfigure}
    \caption{Few-shot \& zero-shot performance.}
    \label{fig:few_zero_shot}
    \vspace{-0.2cm}
\end{figure}

In this experiment, we first trained the prediction model on NYC crime datasets with NYC pre-trained region embeddings, and then evaluated it on CHI crime datasets with CHI pre-trained region embeddings without any fine-tuning. This setup enables a comparative analysis of the zero-shot transferability between our task-learnable prompt and various baselines. Results show in Figure~\ref{fig:zero_shot}.

Compared with baselines without any prompting methods (i.e., HRE, MVURE, and ReCP), models with prompting generally exhibit better zero-shot transferability, suggesting that prompting model can capture task-specific features independent of the city to some extent. Under the same prompting method (prefix prompt), GURP+Pre demonstrates better transferability than HREP, highlighting the superior generalizability of our GURP pre-trained representations. A comparison between GURPP and GURP+Pre reveals that our task-learnable prompting model outperforms the prefix prompt, which can be attributed to the handling of the original region subgraphs within our prompting model.

\subsection{Ablation Studies}
We perform ablation studies on the NYC dataset to evaluate the effectiveness of each individual component of our proposed model. 
\textbf{GURPP} includes both pretraining and task-learnable prompting module. \textbf{GURP} contains only the pretraining module. \textbf{/F}, \textbf{/S}, \textbf{/I}, \textbf{/M} denote the removal of the flow view, spatial proximity view, imagery view, and multi-view fusion from GURP, respectively. \textbf{/K} indicates replacing the knowledge graph embedding initialization with random initialization.

\begin{figure}[htbp]
    \centering
    \vspace{-0.2cm}
    \includegraphics[width=\linewidth]{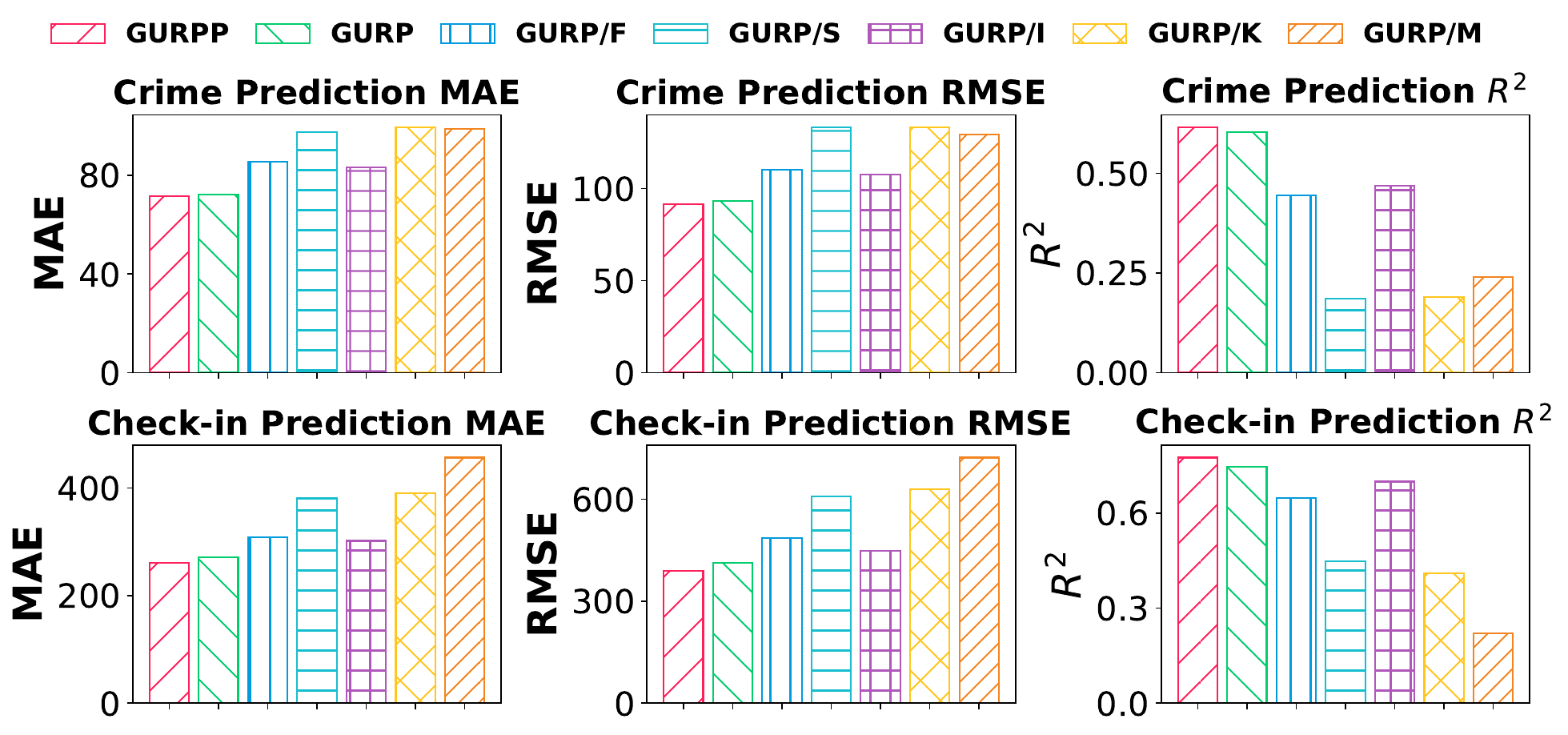}
    \vspace{-0.5cm}
    \caption{Ablation studies.}
    \label{fig:Ablation_all}
    \vspace{-0.3cm}
\end{figure}

Figure \ref{fig:Ablation_all} presents the results. Spatial proximity view and knowledge graph embedding method significantly enhance the model’s performance, lifting it by 151.64\% and 156.67\% of $R^2$, respectively. These components capture the heterogeneous and generic patterns of interactions among entities by elucidating the spatial relationships and semantic knowledge of urban regions. The removal of the multi-view fusion module results in learned representations exerting varying effects across different tasks, which indicates that this module contributes to learning more stable representations. The flow view and the imagery view can also greatly enhance the performance of the model, increasing it by 29.04\% and 21.11\%, respectively. This is because they capture dynamic flow information and visual features of urban regions, respectively, providing the model with a multi-dimensional context that enhances its ability to fully understand and predict urban dynamics. Prompt learning, with an improvement of 3.12\%, effectively elevates model performance by introducing guiding prompts that allow the model to delve deeper into understanding and utilizing task-relevant knowledge.

\vspace{-0.1cm}
\subsection{Scalability Analysis}
\begin{figure}[tbp]
    \centering
    \vspace{-0.3cm}
    \begin{subfigure}[b]{0.235\textwidth}
        \centering
        \includegraphics[width=\textwidth]{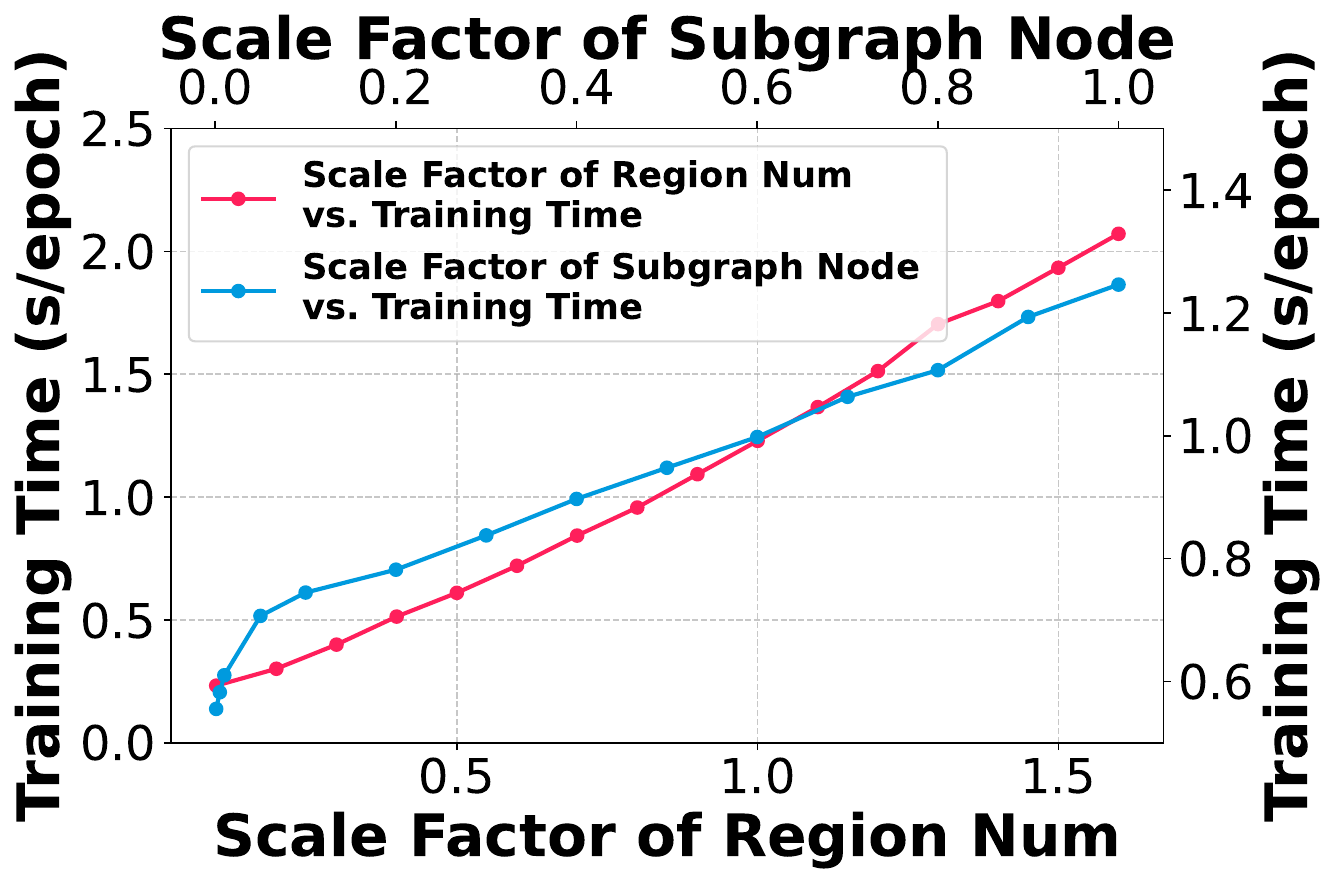}
         \caption{Pre-training Time.}
        \label{fig:sub1}
    \end{subfigure}
    \hfill
    \begin{subfigure}[b]{0.235\textwidth}
        \centering
        \includegraphics[width=\textwidth]{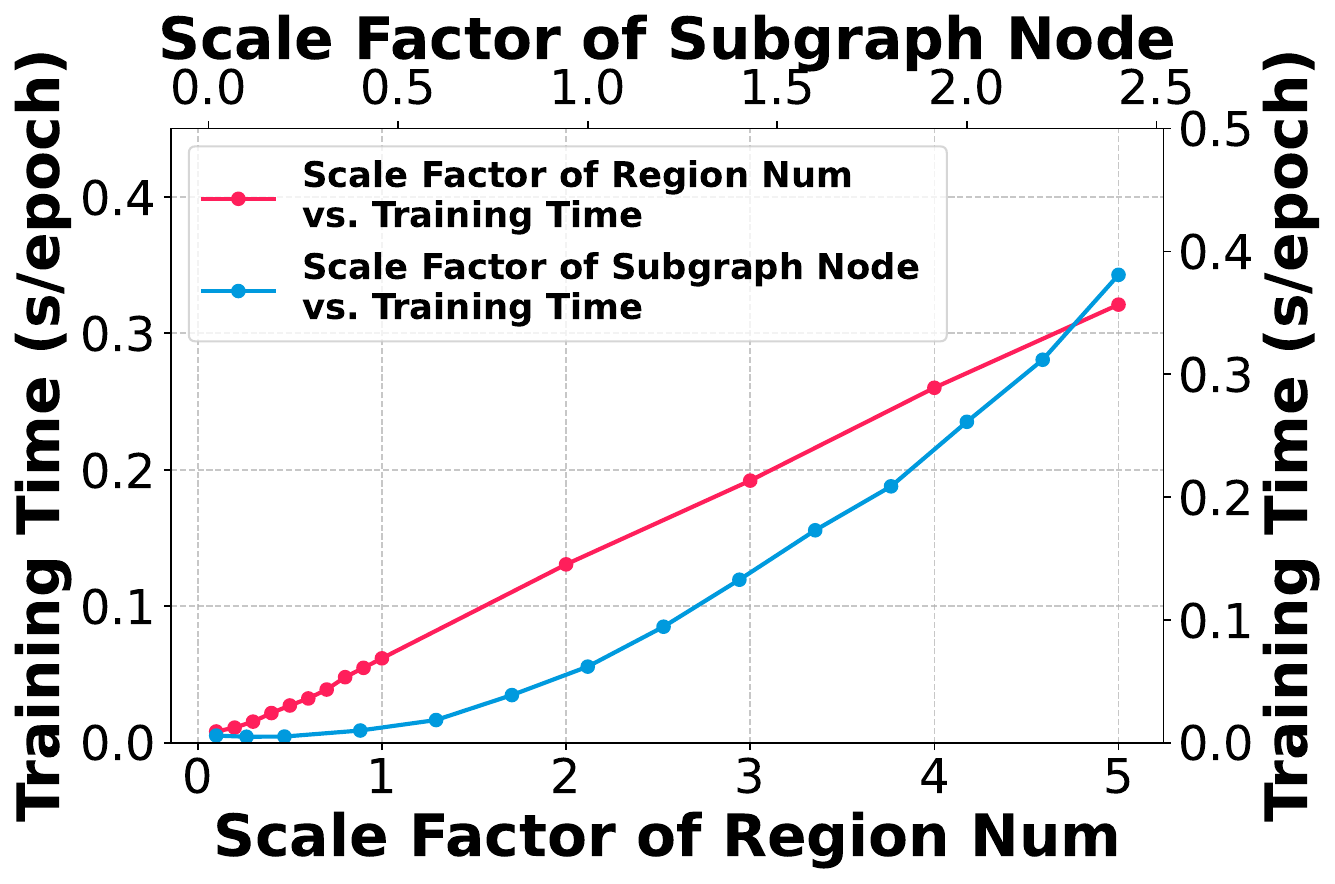}
        \caption{Prompt Tuning Time.}
        \label{fig:sub2}
    \end{subfigure}
    \caption{Impact of data scale on training time.}
    \label{fig:Scalability}
    \vspace{-0.3cm}
\end{figure}

We also validate the scalability of our graph-based method by measuring the pre-training and prompting tuning duration across various scales of regions and region subgraph nodes. Specifically, we use 180 regions in NYC as the base and evaluate the training duration of GURPP over 500 epochs and prompt tuning over 200 epochs for region scale factors of 0.1, 0.2, up to 5 times(1.6 times for pre-training), and evaluate the pretraining duration using 0.05, 0.1, up to 1.0 times the total number of nodes in each region subgraph. For prompt tuning, we conduct evaluations with subgraphs containing 1, 5, up to 120 nodes with the base number 50, which is also the actual experiments settings. The results are averaged over each epoch, shown in Figure~\ref{fig:Scalability}.

As shown in Figure~\ref{fig:sub1}, when the region subgraph size is fixed (GURP uses all nodes and prompt tuning uses 50 nodes per subgraph), the training time for pretraining exhibits a superlinear growth with the number of regions. This is mainly due to the flow feature extraction, where the model incorporates dynamics between all region pairs, whereas prompt tuning scales linearly with the number of regions. When the number of regions is fixed, the pretraining duration shows a sublinear relationship with the number of nodes in the regional subgraph, ensuring the scalability of GURP. Due to the random walk kernel, the training time for the task-learnable prompt scales superlinearly with the number of nodes. We set the number of nodes to 50, maintaining acceptable training efficiency. 

\vspace{-0.1cm}
\subsection{Case Study}
To validate the superiority of GURPP, we conduct a case study to predict the top 10 regions with the highest crime rates in NYC. The study compares our manually designed prompt, $\text{GURPP}_M$-P4, against large the language models (LLMs), including Deepseek R1 (www.deepseek.com) and Gemini 2 Flash (deepmind. google/technologies/gemini/flash/). Detailed descriptions are in Appendix~\ref{app:llmp}.

As shown in the Figure~\ref{fig:case_study_visualization}, GURPP is able to predict crime regions (the regions in the upper half of the figure) but LLMs fail. This indicates that, compared to language prompts, which directly describe the task to induce relevant knowledge in large language models, GURPP explicitly incorporates task-specific knowledge into the representation using graphs, leading to better performance.

\begin{figure}[htbp]
    \centering
    \vspace{-0.3cm}
    \includegraphics[width=0.95\linewidth, height=4.2cm]{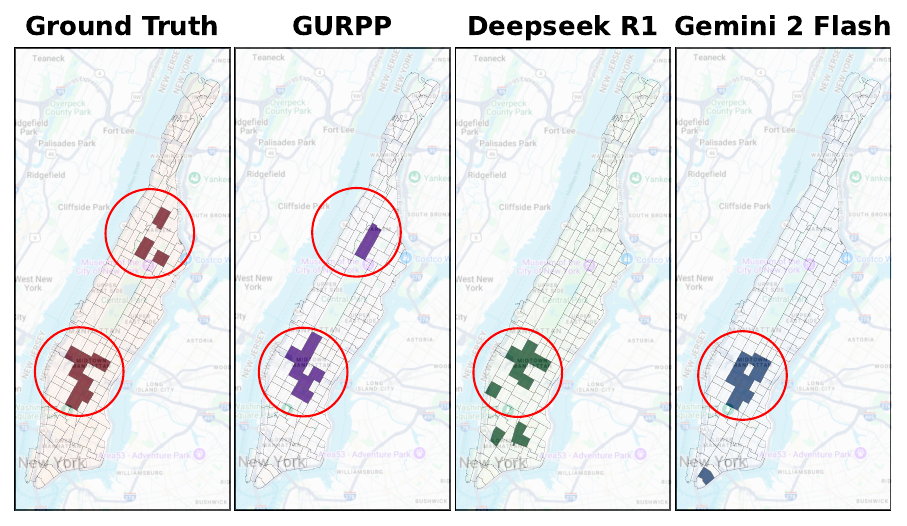}
    \vspace{-0.1cm}
    \caption{Comparison with language prompt. Shaded regions represent potential crime regions.}
    \label{fig:case_study_visualization}
    \vspace{-0.4cm}
\end{figure}
\section{Related Work}

\noindent
\textbf{Urban Region Embedding.}
Urban region representation learning aims to extract complex patterns and relationships within urban regions and embed them into a representation space. Existing research can be broadly classified into two categories: task-specific supervised region representation learning~\cite{bai2019spatio,yang2022classifying,zhou2023predicting} and general unsupervised region representation learning~\cite{li2023urban,zhang2022region,li2024urban,yao2018representing,wu2022multi,jenkins2019unsupervised,zhang2021multi,sun2024urban,xu2024cgap}. These works, focusing more on exploring multi-view fusion methods to generate representation based on predefined region statistics features with limitations in capturing detailed spatial layout and semantics. Moreover, the lack of fine-grained characterization of region features makes it challenging to further adapt the representation to the specificities of downstream tasks. We address the limitations by using an urban region graph and subgraph-centric region embeddings. Although Zhao {\it et al.}~\cite{Zhao2023Learning} employ a graph-based approach to model urban regions at a fine-grained level, their framework is not specifically designed for urban-region pre-training and prompting tasks.

\noindent
\textbf{Pre-training and Prompting.} The Pre-training and prompting paradigms have achieved significant success in NLP~\cite{liu2023pre, roberts2020much}, CV~\cite{yoo2023improving, zhou2022learning, zhou2022conditional}, and graph learning~\cite{sun2023all, yu2024multigprompt, ma2024hetgpt, yu2024hgprompt, liu2023graphprompt, yang2024graphpro,Jiabin2025HiGPT}.
Many works in urban computing have also made attempts in this regard \cite{xu2023urban,zhang2024towards,balsebre2023city,yuan2024unist,yan2024urbanclip,xiao2024refound, li2024urbangpt}. For instance, Yuan \etal~\cite{yuan2024unist} design an urban spatio-temporal prediction pre-train and prompt model. Yan \etal~\cite{yan2024urbanclip} introduce LLMs to describe urban imagery features and unify different prediction tasks through textual prompts. Although the urban tasks they explored differ from ours, their work can inspire ours to integrate LLMs to enhance fine-grained characterization capabilities.
HREP~\cite{zhou2023heterogeneous} introduces prefix prompt to region embedding, but it lacks explicit extraction of task-adaptation. In this way, we propose both manually-designed and task-learnable prompts built on the graph-based pre-train model to address this issue. The categorization of related work is listed in Table~\ref{tab:related-work}.

\begin{table}[ht]
  \small
  \caption{Cats.of related works on urban region embedding.}
  \label{tab:related-work}
  \vspace{-0.2cm}
  \begin{tabular}{|c|c|c|}
    \hline
    \textbf{Category} & \textbf{Non-Graph-based} & \textbf{Graph-based} \\ 
    \hline
    \textbf{Pretraining} &\cite{yan2024urbanclip, zhang2022region,li2023urban,li2024urban, xiao2024refound} & \cite{fu2019efficient,zhang2021multi,wu2022multi,zhou2023heterogeneous}, GURPP (Ours) \\
   \hline
    \textbf{Prompting} & \cite{yan2024urbanclip, zhou2023heterogeneous}& GURPP (Ours) \\
   \hline
  \end{tabular}
  \vspace{-0.2cm}
\end{table}
\vspace{-0.2cm}
\section{Conclusion and Future Works}
In this paper, we introduce the urban region graph and propose a graph-based pre-training and prompting framework for urban region representation learning, balancing urban generality and task specificity. We develop a subgraph-centric pre-training model to capture general urban knowledge and two prompt methods to incorporate task-specific insights. Extensive experiments demonstrate the effectiveness of our approach. Future work may explore: 1) Integrating textual data and LLMs to uncover more universal urban knowledge; 2) Developing arbitrary region representations for flexible urban modeling~\cite{Sun2025FlexiReg}; 3) Assessing the transferability of urban representations across cities to enhance general applicability.

\bibliographystyle{ACM-Reference-Format}
\bibliography{reference}
\appendix
\section{Appendix}
\subsection{Urban Region Graph Construction}\label{app:graph construct}
\begin{figure}[htbp]
    \vspace{-0.2cm}
    \centering
    \includegraphics[width=0.6\linewidth]{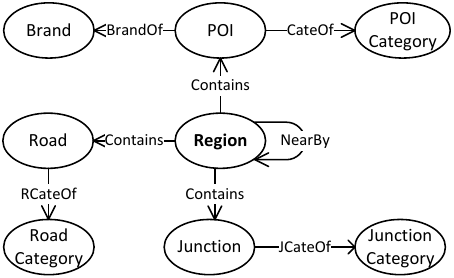}
    \vspace{-0.2cm}
    \caption{Schema of urban region graph.}
    \label{fig:schema of urg}
    \vspace{-0.2cm}
\end{figure}

Here we provide more details for the urban region graph construction, with the schema shown in Figure~\ref{fig:schema of urg}. We define eight node types shared across regions and cities: {\it region, POI, POI category, brand, road, road category, junction, and junction category} in total, comprising the node type set $T_V$, and six edge types to form the edge type set $T_E$. These include the \textit{NearBy} relationship, which connects region entities based on their geographical adjacency; the \textit{Contains} relationship, which is based on spatial containment; the \textit{BrandOf} relationship, which links brands to POI entities; and the \textit{CateOf}, \textit{JCateOf}, and \textit{RCateOf} relationships, which are based on the category information of the entities.

\subsection{Region Imagery Encoding}\label{app:imagery}
We provide details on urban imagery encoding. ResNet~\cite{he2016deep} is utilized to extract visual features for each region image, represented as $\vec{a}_k=\text{ResNet}(a_k)$. A comprehensive visual representation for the region is obtained by averaging features from multiple images: $\vec{h}_r^{img}=\text{Linear}(\frac{1}{m}\sum_{k=1}^{m}\vec{a}_k).$ We then adopt the contrastive loss $\mathcal{L}_{img}$ to align the visual representation with the subgraph-centric embedding.

\vspace{-0.4cm}
\begin{equation*}
\begin{aligned}
    \mathcal{L}_{img} &= \sum_{r \in \mathcal{R}} \Big[
    -\log\exp\left(\frac{\vec{h}_r \cdot \vec{h}_r^{img}}{\tau}\right) \\
    &+ \log\left(\exp\left(\frac{\vec{h}_r \cdot \vec{h}_r^{img}}{\tau}\right) +
    \sum_{j=1}^{B-1} \exp\left(\frac{\vec{h}_r \cdot \vec{h}_j}{\tau}\right)
    \right)
    \Big],
\end{aligned}
\end{equation*}
where, $B$ represents batch size and $\tau$ is the temperature parameter.

\subsection{Mobility Pattern Encoding}\label{app:flow}
We provide a detailed description for mobility pattern encoding. We first reconstruct the departure and arrival flow distribution. Let $m_{ij}$ represent the number of trips from region $r_i$ to region $r_j$. The source and destination urban flow distributions are computed as $P^{{sc}}(r_j | r_i) = \frac{m_{ij}}{\sum_{k=1}^{N} m_{ik}}$ and $P^{{dt}}(r_i | r_j) = \frac{m_{ij}}{\sum_{k=1}^{N} m_{kj}}$, respectively. We then reconstruct the source and destination distributions $\hat{P}^{{sc}}(r_j | r_i)$ and $\hat{P}^{{dt}}(r_i | r_j)$ using the encoded inflow and outflow features:
\[
\hat{P}^{{sc}}(r_j | r_i) = \frac{\exp((\vec{h}_i^{{sc}})^\top \vec{h}_j^{{dt}})}{\sum_k \exp((\vec{h}_i^{{sc}})^\top \vec{h}_k^{{dt}})},
\hat{P}^{{dt}}(r_i | r_j) = \frac{\exp((\vec{h}_j^{{dt}})^\top \vec{h}_i^{{sc}})}{\sum_k \exp((\vec{h}_j^{{dt}})^\top \vec{h}_k^{{sc}})}.
\]

Finally, we learn the inflow and outflow embeddings by optimizing the following loss function:
\begin{equation*}
\mathcal{L}_{flow} = \sum_{(r_i,r_j)} - P^{sc}(r_j | r_i) \log \hat{P}^{sc}(r_j|r_i) - P^{dt}(r_i | r_j) \log \hat{P}^{dt}(r_i|r_j),  
\end{equation*}
which aids in learning inflow and outflow embeddings by comparing true and reconstructed distributions.

\subsection{Attribute Graph Kernel Function}\label{app:kernel}
We introduce how attribute graph kernel function $K_{p}(\cdot,\cdot)$ works. We choose a $P$-step random walk kernel~\cite{feng2022kergnns}~as graph kernel function $K(\cdot,\cdot)$ to measure the similarity between graphs, which counts the number of common walks in both graphs within $P$ steps. Given $G_1=(V_1, E_1),~G_2=(V_2,E_2)$, where $V$ and $E$ are the set of nodes and edges, we initially define the direct product graph $G_{\times}$ \cite{feng2022kergnns}: 
$
G_{\times} = G_1\times G_2 = (V_\times, E_\times),
$
where $V_\times=\{(v_1,v_2):v_1\in V_1, v_2\in V_2\}$, $E_\times=\{\{(v_1,v_2),(u_1,u_2)\}:\{v_1,u_1\}\in E_1 \land \{v_2,u_2\}\in E_2\}$. Following the theorem \cite{vishwanathan2010graph} that a single random walk on the direct product graph $G_\times$ is equivalent to random walk on the two graphs $G_1,~G_2$ at the same time, the $P$-step random walk kernel is computed by:
\[
    K(G_1,G_2) = \sum_{p=0}^PK_p(G_1,G_2) = \sum_{p=0}^P\lambda_p\sum_{i,j=1}^{|V_\times|}[A_{\times}^p]_{ij},
\]
where $A_\times$ is the adjacency matrix of the direct product graph, $\lambda_p$ is the sequence of weights and the $(i,j)$ item of $A_\times^p$ represents the $P$-length common walks between the $i$-th and $j$-th nodes. 

Extending the above definition to the scenario of graphs with node attributes, we use $\mathbf{X}\in\mathbb{R}^{n\times d}$ to denote the attribute matrix of a graph with $n$ nodes and attribute dimension $d$. The node attribute matrices of $G_1,G_2$ are denoted by $\mathbf{X}_1\in\mathbb{R}^{n_1\times d}, \mathbf{X}_2\in\mathbb{R}^{n_2\times d}$. Thus, for the graph $G_\times =G_1\times G_2$, the node attribute matrix is $\mathbf{S}=\mathbf{X}_1\mathbf{X}_2^\top, \mathbf{S}\in\mathbb{R}^{n_1\times n_2}$, where the item $(i, j)$ encodes the similarity between the $i$-th node in $G_1$ and the $j$-th node in $G_2$. We flattened $\mathbf{S}$ to $\mathbf{s}\in\mathbb{R}^{n_1n_2}$ to obtain the formula for the random walk kernel:
\[
    K_p(G_1,G_2)=\sum_{i,j=1}^{|V_\times|}\mathbf{s}_i
    \mathbf{s}_j[A_\times^p]_{ij}=\mathbf{s}^\top A^p_\times \mathbf{s}.
\]

\subsection{Experiment Details}\label{app:experiment details}
\subsubsection{Dataset Statistics and Sources.}\label{app:dataset}
Statistics and sources of the datasets are shown in Table \ref{tab:dataset}. The region division is based on census tracts and the urban imagery are 256 x 256 pixels with a resolution of 4.7 meters. For carbon emissions data provided in 1x1 $km$ grid cells, we map them to our regional divisions. If a region covers more than one grid cell, the emissions from all relevant cells are summed. For the income and education data, each region is assigned a specific value representing the average income or number of people with an education for that region.

\begin{table}[htbp]
  \small
  \caption{Dataset Statistics and Sources.}
  \vspace{-0.4cm}
  \label{tab:dataset}
  \begin{tabular}{c|cc|c}
    \toprule
    Data & NYC & CHI & Source\\
    \midrule
    Region & 180 & 869 & Census Bureau~\cite{Census} \\
    \midrule
    POI & 143,445 & 112,989 & OSM~\cite{OpenStreetMap}, Safegraph~\cite{Safegraph}\\
    Imagery & 7,160 & - & ArcGIS~\cite{Arcgis, liu2023knowledge}\\
    Taxi trip & 12,684,353 & 3,889,032 & NYCOD~\cite{NYCOD}, CHIDP~\cite{CDP}\\
    Road & 110,919 & 139,850 & OSM~\cite{OpenStreetMap}\\
    Junction & 62,437 & 66,065 & OSM~\cite{OpenStreetMap}\\
    \midrule
    Crime & 35,335 & 274,733 & NYCOD~\cite{NYCOD}, CHIDP~\cite{CDP}\\
    Check-in & 106,902 & - & Foursquare~\cite{Foursquare}\\
    Crash & - & 150,535 & CHIDP~\cite{CDP}\\
    Carbon & 1x1 $km$ & 1x1 $km$ & ODIAC~\cite{oda2015odiac} \\
    Income & 180 & - & NYCOD~\cite{NYCOD} \\
    Education & - & 869 & CHIDP~\cite{CDP} \\
    \bottomrule
  \end{tabular}
  \vspace{-0.3cm}
\end{table}

\subsubsection{Baselines Descriptions.}\label{app:baselines}
Here we provide more details about the three types of baseline methods.

\noindent I. Knowledge graph embedding methods:
\begin{mbi}
    \vspace{-0.1cm}
    \item \textbf{TransR} \cite{lin2015learning} projects entities and relationships into relation-specific spaces. \textbf{TransR-N} uses the region node embedding as the representation. \textbf{TransR-G} averages embeddings of all nodes in the region subgraph.
    \vspace{-0.1cm}
\end{mbi}
\noindent II. Graph embedding methods:
\begin{mbi}
    \vspace{-0.1cm}
    \item \textbf{node2vec} \cite{grover2016node2vec} uses random walks to generate node sequences and learns representations with the Skip-Gram model.
    \item \textbf{GAE} \cite{kipf2016variational} has an encoder for node representations and a decoder for reconstructing the adjacency matrix.
    \vspace{-0.1cm}
\end{mbi}
\noindent III. State-of-the-art urban region embedding methods:
\begin{mbi}
    \vspace{-0.1cm}
    \item \textbf{MVURE} \cite{zhang2021multi} constructs multi-view data based on human mobility and region attributes for region representation.
    \item \textbf{MGFN} \cite{wu2022multi} learns relationships among human mobility patterns through cross-attention.
    \item \textbf{HREP} \cite{zhou2023heterogeneous} creates a multi-relational graph and a relation-aware GCN with a non-graph prefix prompt. \textbf{HRE} is a variant without the prompt module.
    \item \textbf{ReCP} \cite{li2024urban} learns region representation through a multi-view fusion that captures unique and consistent information.
    \vspace{-0.1cm}
\end{mbi}

\subsubsection{Implementation.}\label{app:implementation}
We use five-fold cross-validation to evaluate GURPP and baselines on multiple tasks across both cities. In GURPP, the region embedding dimension is set to 144 \cite{zhou2023heterogeneous}. The HGT layer and head numbers are 2 and 4, respectively; triplet margin $\delta = 2$; fusion loss weight $\mu = 0.01$; learning rate is 0.001; and weight decay is 1e-6. In the task-learnable prompt module, we set the random walk kernel layers to 2, prompt graph sizes to 6 and 8, and region subgraph node size to 50. Due to the lack of satellite image datasets for CHI, we removed the image encoding module and adjusted the fusion layer dimensions accordingly.

Our model is implemented with PyTorch 1.8 and DGL 2.2.1+cu118 on a machine with an NVIDIA RTX 3090 GPU, 64 GB RAM, and a 3.50 GHz CPU. Scalability experiments use an NVIDIA H100 80GB GPU, 1 TB RAM, and a 3.2 GHz CPU.

\subsubsection{Comparison with Language Prompt.}\label{app:llmp}
We conduct a comparison with LLMs, including Deepseek R1, Gemini 2 Flash, in predicting the top 10 regions with the highest crime probabilities in NYC. The specific variant of GURPP used in this comparison is $\text{GURPP}_M$-P4. We provide each language model with detailed descriptions of the 180 regions in NYC, including region IDs and information about POIs. The prompt is as follows.

\begin{tcolorbox}[colframe=black, colback=gray!10, left=2mm, right=2mm, top=1mm, bottom=1mm]
    \small
    I will provide information on the division of 180 regions in Manhattan, New York, with each region including a region ID and detailed description. Please analyze the potential crime risk of each area based on their descriptions, rank them according to the likelihood of criminal incidents from highest to lowest, and list the top 10 highest-risk regions. The regional information is formatted as: \{\textit{mhtr0: mhtr0 is located in Manhattan, encompassing 1318 POIs, with the main types being Restaurants and Other Eating Places, Clothing Stores ...}\} ...
\end{tcolorbox}

\end{document}